\documentclass{article}

\PassOptionsToPackage{numbers}{natbib}
\usepackage[final]{ap_2021}
\usepackage[utf8]{inputenc} % allow utf-8 input
\usepackage[T1]{fontenc}    % use 8-bit T1 fonts
\usepackage{hyperref}       % hyperlinks
\usepackage{url}            % simple URL typesetting
\usepackage{booktabs}       % professional-quality tables
\usepackage{amsfonts}       % blackboard math symbols
\usepackage{nicefrac}       % compact symbols for 1/2, etc.
\usepackage{microtype}      % microtypography
\usepackage{nccmath}
\usepackage{graphicx}
\usepackage{caption}
\usepackage{subcaption}
\usepackage{listings}
\usepackage{longtable}
\usepackage{ulem}
\usepackage{xcolor}
\usepackage{soul}
\usepackage{vcell}

\title{Proto: A Neural Cocktail for Generating Appealing Conversations}

\author{
	Sougata Saha$^{*}$, Souvik Das\thanks{Equal Contribution and team leads. Department of Computer Science \& Engineering.}, Elizabeth Soper$^\dagger$, Erin Pacquetet\thanks{Authors are in alphabetical order. Department of Linguistics.}, Rohini K. Srihari\thanks{Faculty advisor. Department of Computer Science \& Engineering.} \\
	Department of Computer Science \& Engineering, Department of Linguistics\\
	University at Buffalo\\
	Buffalo, NY 14260 \\
	\texttt{\{sougatas, souvikda, esoper, erinmorr, rohini\}@buffalo.edu} \\
	\and 
	
}

\begin{document}

\maketitle

\begin{abstract}
In this paper, we present our Alexa Prize Grand Challenge 4 socialbot: Proto. Leveraging diverse sources of world knowledge, and powered by a suite of neural and rule-based natural language understanding modules, state-of-the-art neural generators, novel state-based deterministic generators, an ensemble of neural re-rankers, a robust post-processing algorithm, and an efficient overall conversation strategy, Proto strives to be able to converse coherently about a diverse range of topics of interest to humans, and provide a memorable experience to the user. In this paper we dissect and analyze the different components and conversation strategies implemented by our socialbot, which enables us to generate colloquial, empathetic, engaging, self-rectifying, factually correct, and on-topic response, which has helped us achieve consistent scores throughout the competition.

 \end{abstract}

\section{Introduction}
Expressing thoughts and interacting with one another using dialogue is instinctive for humans. Hence, creating an artificially intelligent socialbot that can converse with a human, at par with human-level conversational capability, is very difficult. Proto is our attempt to create a socialbot that exhibits human-level conversation competency, and engages in conversation with humans, on any permissible topic. With a state-of-the-art ensemble of neural response generators, which are powered by large scale knowledge bases like Wikidata, and leveraging other latest research in natural language processing and conversational AI, Proto is an end-to-end conversational system which can not only understand and generate responses, but also drive the conversation deeper, be it any topic. This paper introduces and describes in detail how our Alexa Prize 2020 socialbot is able to achieve consistently high scores in conversations with real users.

A socialbot is generally defined as an artificially intelligent agent, that can participate in an open domain conversation with a human user on any topic, much like talking to a stranger at a bar. Unlike task or goal-oriented systems, where the bot aims to perform certain actions like successful ticket reservation, customer support, etc. the unrestricted domain of socialbots, has no well-defined objectives other than engaging in interesting conversations. This introduces novel and exciting challenges, which demand multi-disciplinary solutions. 

While keeping Grice's Maxims in mind, several considerations need to be accounted for, in order to generate an appropriate and succinct response to a user query. We broadly identified and incorporated the following technologies and strategies in our system, in order to handle the wide range of user needs. 
\begin{itemize}
    \item \textbf{Natural Language Understanding:} The ability to understand and extract features (intent, sentiment, entities, etc.) from the user's query. In order to generate an appropriate response to a user query, it is imperative to understand and extract features from the query, which can be acted upon by the response generators.
    \item \textbf{Curated Content Selection and Delivery:} The ability to select and incorporate relevant factual content (encyclopedic knowledge) and interesting content (fun facts) in order to generate a response. To be able to generate factual, interactive, and interesting responses, it is important to efficiently collect, process, and store disparate knowledge sources offline, and efficiently retrieve and utilize them with minimum latency online. Leveraging Elasticsearch indexes containing Wikidata and scraped interesting facts, we are able to increase the topic coverage of the bot.
    \item \textbf{Colloquialism:} The ability to generate responses in casual (day-to-day) language. During the competition, we observed that template-based handwritten response generally lack the naturalness and colloquialism which is required by a bot to sound like a human, we also observed that distinguishing natural responses from template-based handwritten responses is a relatively easy task for humans, which can give the perception of the bot not actually understanding the user, and making the user skeptical about the bot. Our ensemble of neural generators trained on diverse human-to-human conversations helps alleviate this problem, and makes the conversation more natural, and increases the humanness of the bot.
    \item \textbf{Remembering User Attributes:} The ability to retain and reuse specific pieces of information shared by the user (example: name, favorites, etc.). Remembering and using personal information generally helps make the conversation more colloquial and cordial. The natural language understanding module extracts slots from the user's query, which contain the user's name and preferences around specific topics (favorite movie, food, etc.), which is used throughout the conversation.
    \item \textbf{Exhibiting Emotions:} The ability to understand user emotions and respond emphatically. As humans are emotional creatures, emotionally adept responses help bridge the barrier between the bot and the user. We noticed that many times the users seek emotional support from the bot and share their personal encounters with the bot, and an empathetic response to such user queries generally leads to a better user experience, thus resulting in better approval ratings. Leveraging a suite of neural generators trained on human-to-human empathetic conversations, complimented by hand-written scenario-based empathetic responses, our bot is able to demonstrate empathy in its responses.
    \item \textbf{Diving Deep:} The ability to dive deep into a topic, and engage in logical exchanges with the user. Human conversations are generally multi-layered in nature. For example, when humans discuss the theme `movies', the initial layers might consist of talking about liking or disliking movies in general, with the further layers discussing the reason for the same, followed by discussing specific movies and specific scenes from the movies which created an impression. Also, in daily conversations, switching to a different, related theme on encountering certain trigger words or phrases is common. For example, while discussing cars, switching the topic to the movie `Fast and Furious' seems logical and natural. We tried to mimic this behavior of diving deep into conversations and naturally transitioning to related themes in our bot, through our deterministic and neural generators. A robust retrieval and re-ranking-based prompt selection module helps us dive deeper into scenarios where the generators are incapable of advancing the conversation, and also enables the bot to transition to a logical next theme, based on certain words or phrases in the context of the current discussion. We found this to be one of the most difficult criteria to tackle, but is crucial for achieving truly human-like conversation. 
    \item \textbf{Maneuverability:} The ease of navigating and eliciting response or reaction from the bot. Just like driving a car, the user should be able to steer the bot with ease and should be able to evoke a reaction or response from the bot using dialogue. For example, they should be able to change the current topic of discussion, get a response to queries (empathetic, opinions, etc.), go back to a previous topic of discussion, suggest topics or entities to discuss, express dissatisfaction, and end the conversation when desired. Our suite of natural language understanding modules helps us understand the intent of the user better, which aids in generating a desirable response. Throughout the competition, we noticed a positive correlation between high user ratings and increased maneuverability of the bot. 
    \item \textbf{Programmatic Change of Control:} The ability to change the narrative of the conversation by switching the control of the conversation between the user and the bot, as and when required. Generally in human-to-human conversations, both actors are responsible for advancing the conversation. However, in human-to-bot conversations, we noticed that users, in general, don't help much in progressing the conversation, and resort to bland and short responses. Hence, the onus is on the bot to advance the conversation, which is best done by asking questions to the user. However, a drawback of this strategy is that the user experience might be hurt due to questions in every turn. Hence, we not only implement strategies to prevent prompts in every turn thus providing control to the user but also encourage the user to introduce new themes and entities in the conversation, in a controlled way.
    \item \textbf{Self-Rectification:} The ability to recover from an undesired state in the conversation. Sometimes the bot might misunderstand the intent of the user, hence enter a state which is inappropriate at that point in time, and generate misleading responses. In the future turns, this problem might cascade, and take the conversation in a different direction altogether. To recover from these states, we actively ask feedback from the user every \textit{k} turns, to check if they are interested in continuing with the current theme or want to switch to a different topic. Our natural language understanding module is also able to detect dissatisfaction, negative intent, and complaints from the user, which we can then respond to with an appropriate message, and transition the discussion to a different theme.
    
\end{itemize}
Having given a high-level overview of the features of a successful socialbot, in the following sections we discuss how we have realized these features in detail, and then analyze the impact of the features on users' experience with the bot.

% \section{System architecture and Overview}
% \label{gen_inst}

\section{End-to-End Flow: Walk-through of response generation}
\label{headings}
\begin{figure}[h]
    \centering
    \includegraphics[scale=0.4]{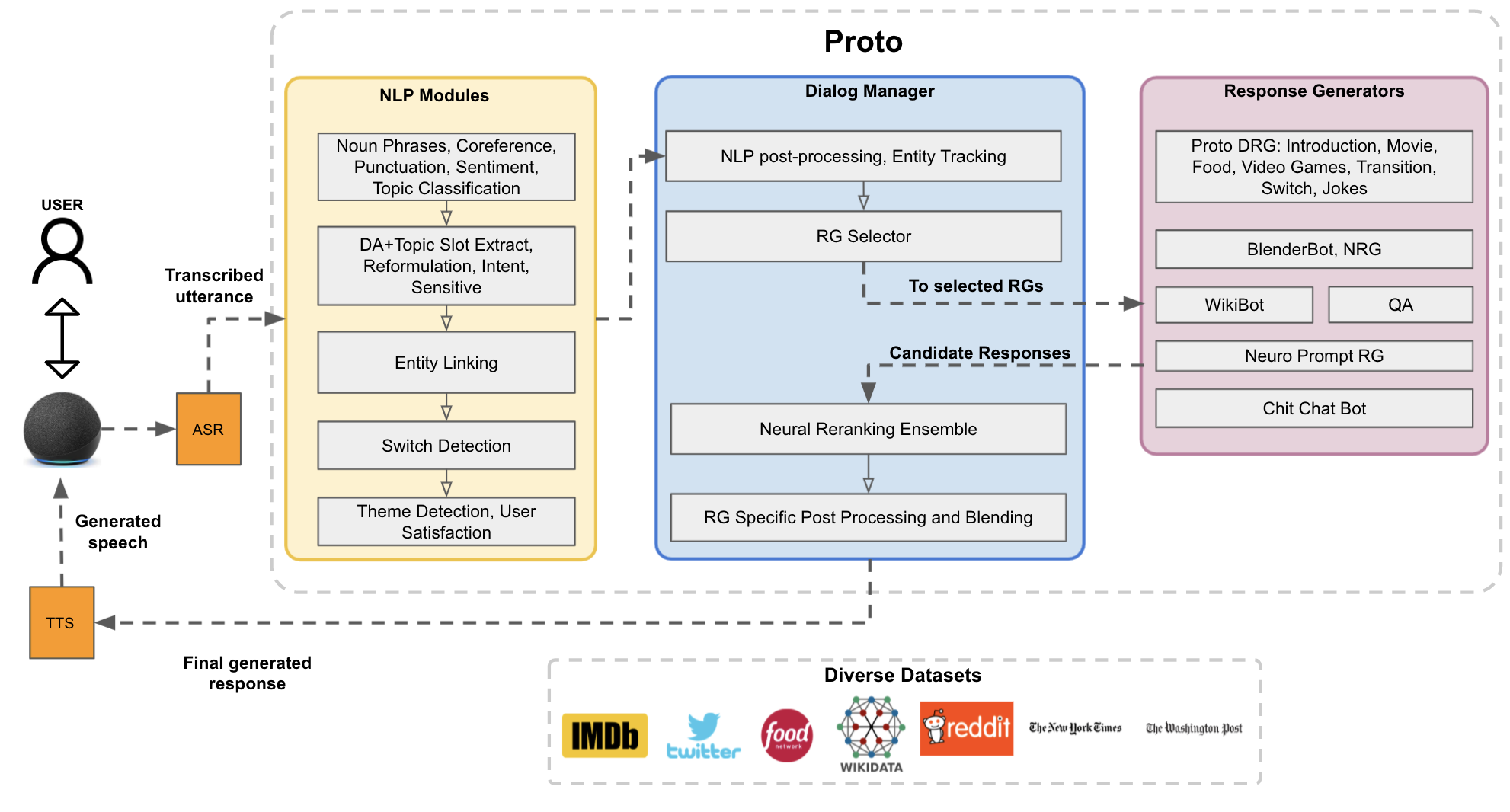}
    \caption{Proto System Architecture }
    \label{fig:proto_arch}
\end{figure}

Figure \ref{fig:proto_arch} illustrates our end-to-end architecture at a high level. The journey of a user query inside our bot starts when the Amazon Speech-to-Text model converts user speech into text and passes the query to our bot. Upon receiving a text query, the dialog manager invokes a suite of Natural Language Understanding (NLU) modules, which extract diverse features like user sentiment, entity mentions, conversation topic \& theme, user current satisfaction, user intent, diverse slot values .etc. Post feature extraction, the dialogue manager links and tracks the entity mentions (if any) to Wikidata, and invokes an appropriate set of response generators in parallel, in order to generate a set of candidate responses to the user query. Once the candidate responses are generated, an ensemble of neural re-rankers utilizes the current query and the conversation context for ranking and selecting the best response. A post-processing module enhances the best response by adding prompts, salutations .etc, or removing contradictions, excess punctuation, phrases .etc. The response from the post-processing module is returned to Amazon's Text-to-Speech model, which generates and delivers the audio response to the user. The dialog manager saves the state of the current conversation in a DynamoDB state table, which is reused in the next turn. In the next sections, we explain each of the components in detail, which enables us to generate an accurate response in minimum latency.\\

\section{Natural language understanding modules}
\label{others}
The natural language understanding modules extract several features from the utterance, which are essential for the different downstream modules. The outputs of the NLU modules are stored in the state manager of the conversation, which gets written into DynamoDB.

\subsection{Query Punctuation, Named Entity Recognition \& Sentiment Detection:} Since the text provided by the ASR module does not contain any punctuation, we leverage the BERT-based punctuation model present in the Cobot environment for re-punctuating the user query.

Detecting the entities from the user text is a crucial task. Accurate entity extraction not only enables our bot to stick to the topic the user is interested in but also helps us retrieve relevant factual information regarding the entity of interest. We leverage the BERT-based NER model extended in Cobot for extracting entities from the user query.

Along with determining the entities, it is important to understand the sentiment of the user at any given point of the conversation. For our purpose, we leverage the lightweight VADER (Valence Aware Dictionary and Sentiment Reasoner) sentiment analyzer and restrict the sentiment classes to be positive, negative, or neutral, along with the intensity of the sentiment, ranging between -1 to 1. Efficient sentiment detection enables us to generate emotionally appropriate responses and enhances maneuverability across topics, thus providing an engaging experience to the user. 

\subsection{User Intent Detection}
\begin{table}[t]
  \centering
  \resizebox{12cm}{!}{%
  \begin{tabular}{lll} %p{4cm}p{2cm}
    \toprule
    \cmidrule{1-2}
    \textbf{Name}    & \textbf{Example}     & \textbf{\# Training Examples}\\
    \midrule
    acknowledgement & \textit{Yes}  & 165   \\
    rejection    &  \textit{No}  & 92 \\
    clarification   & \textit{What?, I didn’t understand that.}  & 80\\
    state\_personal\_fact & \textit{I have a dog}. & 1093\\
    state\_knowledge\_fact & \textit{Water boils at 100 degrees celsius}. & 1111\\
    state\_opinion & \textit{I like this film.} & 1797\\
    request\_personal\_fact & \textit{Do you have a dog?} & 571\\
    request\_knowledge\_fact & \textit{How old is George Clooney?} & 297\\
    request\_opinion & \textit{Do you like comedies?} & 298\\
    topic\_suggestion & \textit{let’s talk about movies.} &  31 \\
    general\_chat & \textit{How are you?} & 486 \\
    other & Anything other than the above classes & 171\\
    \bottomrule
  \end{tabular}%
}
  \caption{Types of User Intent}
  \label{intent-table}
\end{table}

Human conversations exhibit certain patterns. For example, when someone asks a personal question, the response is generally a personal statement or an opinion. When a piece of information is requested, the response is generally a factual statement. In order to leverage such patterns, it is imperative to understand the intent of a user query. Analyzing existing conversation datasets, we define 12 intent categories that commonly exist in a conversation. Table \ref{intent-table} details the defined intent classes. We annotated a subset of Topical Chat dataset \cite{gopalakrishnan2019topical}, Chit-Chat dataset \cite{myers2020conversational}, and ConvAI-2 dataset \cite{convai2} to create training examples for a BERT-based \cite{devlin2019bert} intent classifier, which is hosted outside of the Cobot environment, and accessible using API calls. Table \ref{tab:intent} details the model's performance on the test set.

\begin{table}[h]
\centering
    \begin{minipage}{.45\linewidth}
        \begin{tabular}{lll}
        \hline
        \textbf{Model}  & \textbf{F1}   & \textbf{Accuracy} \\ \hline
        bert-base-uncased & 71.48 & 71.7     \\ \hline
        \end{tabular}
        \caption{BERT-based intent classifier met-\\rics on test set}
        \label{tab:intent}
    \end{minipage}
    \begin{minipage}{.45\linewidth}
        \begin{tabular}{lll}
        \hline
        \textbf{Model}  & \textbf{F1}   & \textbf{Accuracy} \\ \hline
        bert-base-uncased & 97.78 & 97.74    \\ \hline
        \end{tabular}
        \caption{BERT-based sensitive classifier metrics on test set}
        \label{tab:sensitive}
    \end{minipage}
\end{table}

% \begin{table}[h]
% \centering
% \begin{tabular}{lll}
% \hline
% \textbf{Model}  & \textbf{F1}   & \textbf{Accuracy} \\ \hline
% bert-base-uncased & 65.4 & 65.3     \\ \hline
% \end{tabular}
% \caption{BERT-based intent classifier metrics on test set}
% \label{tab:intent}
% \end{table}

Unfortunately, many times users try to elicit a reaction from the bot by using foul and sensitive language. In order to detect such queries, we implement an iterative approach. In the first iteration, we leverage the high precision, keyword-based sensitive classifier developed by Stanford Chirpy Cardinal \cite{paranjape2020neural}. In order to increase the recall, in the second iteration, we leverage a BERT-based sensitive query detection classifier, which is only invoked if the text is not classified as sensitive in the first iteration. For training the BERT-based sensitive query classifier, we annotated a subset of the CAPC dataset (shared by Amazon), and the Bot-Adversarial Dialogue dataset \cite{xu2020recipes}. The combined dataset contains 24,250 non sensitive examples and 21,869 sensitive examples. Table \ref{tab:sensitive} details the model's performance on the test set.

% \begin{table}[h]
% \centering
% \begin{tabular}{lll}
% \hline
% \textbf{Model}  & \textbf{F1}   & \textbf{Accuracy} \\ \hline
% bert-base-uncased & 97.78 & 97.74    \\ \hline
% \end{tabular}
% \caption{BERT-based sensitive classifier metrics on test set}
% \label{tab:sensitive}
% \end{table}

\subsection{Slot Based Information Extraction}
In order to extract actionable information from a user query, we analyzed a subset of existing conversations with our bot and developed a high-precision regex-based slot extraction module. The module can extract the user's name, age, relationship mentions, liking, disliking, wanting, and intent to switch the discussion to a different topic or entity. The extracted information is generally used by downstream modules, and certain slots like user name, user favorites, etc. are persisted and reused throughout the conversation.

\subsection{Query Reformulation}
Colloquial human conversations are rife with co-reference, making it natural for users to use anaphora while interacting with our socialbot, which makes the task of understanding user intent challenging. Leveraging the conversation context, we implement an ensemble of Spacy neuralcoref \cite{clark-manning-2016-deep, clark-manning-2016-improving}, and Allen AI co-reference resolution \cite{Lee2017EndtoendNC} to reformulate the latest user query and make it more adequate with information. We use Allen AI's co-reference resolution as the primary resolver but resort to Spacy if it fails to resolve the co-references.

\subsection{Entity Linker}
To understand precisely which entities a user is talking about in a conversation, Entity Linking plays an important role. This module is extensively used in several areas in our system like response generation, topic understanding, entity tracking, etc. To make the system scalable the Entity Linking engine is decoupled from the Cobot framework and hosted in an EC2 instance. Along with our key changes in scoring strategy to address the conversational context and rarity of the spoken entity,our approach was inspired in large part by last year's Chirpy Cardinal's open-source release\cite{paranjape2020neural}.

We processed a dump of English language Wikipedia. The following information for an entity (referred as $E$) is collected (a)  \textit{pageviews} (number of pageviews in a month) (b) the \textit{anchortext distribution} $P_{anchortext}(a|E)$

The anchortext distribution for an entity $E$ is calculated using the following formula:

\begin{ceqn}
\begin{align}
    P_{anchortext}(a|E) = \frac{fuzzycount(\textrm{links from $a$ to $E$})}{\sum_{a{}'\in A(E)}count(\textrm{links $a{}'$ to $E$})}
\end{align}
\end{ceqn}

Here, we count the fuzzy matches between \textit{anchortexts} (lowercase) with the hyperlinks related to the entity $E$. The python library \textit{fuzzywuzzy} is used to get the token similarity, the similarity threshold is kept at $0.95$. $A(E)$ is a set of all anchortext for the entity $E$. 

The user utterance $u$ is segmented in n-grams where n = 5 and stored into a set $S$. After that, an Elastic Search query is constructed to fetch all entities $E$ which has at least one span $s\in S$. $S$ is used to constitute the search keywords, where stop words are neglected and a boost parameter is provided where the uni-gram frequency of the set element is less than $10$. To link the entity we estimate $P(E|s)$, i.e. likelihood of a span $s$ is referring to an entity $E$, $P(E|s)$ can be estimated using a Bayesian model:

\begin{ceqn}
\begin{align}
    P(E|s)\propto P(E) \times P(s|E)
\end{align}
\end{ceqn}

Here, $P(E)$ is assumed to be equal to \textit{pageviews} and $P(s|E) = P_{anchortext}(s|E)$. Also, we introduce a rarity coefficient $\alpha$ which factors in the rarity of occurrence of an entity and $\theta$ as contextual coefficient which accounts for the context of the entity in a conversation. Therefore, the $score(s, E)$ between the span $s$ and entity $E$ can be defined as:

\begin{ceqn}
\begin{align}
   score(s, E) = \alpha \times \theta \times P(E) \times P_{anchortext}(s|E)
\end{align}
\end{ceqn}

\begin{ceqn}
\begin{align}
    \alpha = \frac{1}{\textrm{unigram frequency($s$)} + 1}
\end{align}
\end{ceqn}

For calculating the contextual coefficient, we concatenate the last two utterances and the current utterance into a string joined by \verb|[SEP]| token, this is referred to as $context$. Also, each entity $E$ is represented by the top three items in the categories field obtained from the Wikipedia dump. The three strings are concatenated into a string joined by \verb|[SEP]| token, this is referred to as $c_{representation}$. Now the context and candidate representations are passed through a cross-encoder which scores the entities based on the context. The raw scores are then normalized and we discard entities that have scores below a threshold. $\theta$ is defined as follows:

\begin{ceqn}
\begin{align}
    \theta  = \textup{NORMALIZE}(CrossEncoder(\textup{context}, c_{representation}))
\end{align}
\end{ceqn}

For ranking the entities we follow the same strategies as used by previous year's Chirpy Cardinal's system. 

\subsection{Entity level Auxiliary Data}
While processing the Wikipedia dump, we also process some auxiliary data for all entities which are used by the response generators: (1) \textbf{Summary sentences: }We are using LexRank\cite{erkan2004lexrank} summarizer to get top-5 sentences from the overview section of an entity, these sentences are treated as facts by the neural generators (2) \textbf{Key-phrases: }We are also extracting top-10 key-phrases from the overview section using Rake\cite{rose2010automatic}. These key phrases are used to contextually switch to a related entity if the response generator hits a roadblock.

\subsection{Current topic \& Theme Detection}
In order to generate a relevant response and guide the conversation better, it is important to understand the topic of a user utterance. However, an utterance can belong to multiple topics. For example, the statement "I was playing Fifa while eating a bag of chips" contains references to both video games and food. Also, there are scenarios where a user tries to unnaturally drift to another topic. For example, when asked "What is your favorite movie ?", a user can respond with "I ate burgers for dinner today. " Although unnatural, we have observed such scenarios, hence making it important to efficiently detect and track the topic. In order to tackle these unnatural topic changes, we introduce the concept of theme, where we constrain a single utterance to belong to only one topic. We employ an ensemble of the Cobot environment's inbuilt topic classifier, entity linker's predicted topic, and a RoBERTa-based classifier \cite{liu2019roberta} trained on the ConCET dataset \cite{ahmadvand2020concet} to detect the most probable topic clusters of the current utterance. The predicted topic cluster from the current query, along with the theme of the previous turn, helps us decide if the current theme of the conversation needs to be changed. The RoBERTa based classifier is trained to classify a sentence into 1 of 17 classes: attraction, celebrities, chitchat, fashion, fitness, food, games, joke, literature, movie, music, news, other, pets animals, sports, tech, and weather. The classifier achieves an accuracy of 0.97, and macro F1 score of 0.94 on a held-out test set. Table \ref{tab:theme_results} details the classification metrics achieved for each of the themes, on the test set.

\begin{table}[]
\centering
\resizebox{10cm}{!}{%
\begin{tabular}{lcccll}
\hline
\textbf{Class} &
  \multicolumn{1}{l}{\textbf{Precision}} &
  \multicolumn{1}{l}{\textbf{Recall}} &
  \multicolumn{1}{l}{\textbf{F1}} &
  \textbf{Support} &
  \textbf{Training examples} \\ \hline
attraction   & 0.95 & 0.95 & 0.95 & 687    & 5,949   \\
celebrities  & 0.89 & 0.89 & 0.89 & 168    & 1,575   \\
chitchat     & 0.99 & 0.99 & 0.99 & 4,331  & 39,401  \\
fashion      & 0.95 & 0.99 & 0.97 & 74     & 658     \\
fitness      & 1.00 & 0.92 & 0.96 & 12     & 102     \\
food         & 0.98 & 0.98 & 0.98 & 334    & 3,205   \\
games        & 0.92 & 0.69 & 0.79 & 16     & 169     \\
joke         & 0.96 & 1.00 & 0.98 & 54     & 486     \\
literature   & 0.99 & 0.99 & 0.99 & 180    & 1,651   \\
movie        & 0.97 & 0.98 & 0.98 & 16,645 & 149,283 \\
music        & 0.99 & 0.98 & 0.98 & 2,963  & 26,636  \\
news         & 0.95 & 0.94 & 0.94 & 6,339  & 57,807  \\
other        & 0.93 & 0.98 & 0.95 & 645    & 5,361   \\
pets animals & 0.97 & 0.92 & 0.95 & 250    & 1,939   \\
sports       & 0.96 & 0.94 & 0.95 & 2,159  & 18,931  \\
tech         & 0.97 & 0.96 & 0.97 & 2,213  & 20,469  \\
weather      & 0.90 & 0.82 & 0.86 & 66     & 602     \\ \hline
\end{tabular}%
}
\caption{Theme classifier performance on the test dataset. }
\label{tab:theme_results}
\end{table}

\subsection{User Satisfaction Detection}
Detecting the level of user satisfaction with the current theme and bot's responses, and acting accordingly is crucial for delivering a good user experience. Inspired by Chirpy Cardinal's satisfaction detection module, and complemented with our own suite of detectors, we implement a set of regex-based detectors, which can detect user complaints and level of engagement, and help generate better responses, or redirect the conversation to a more relevant theme if needed. 

\subsection{Fact Retrieval}
Interesting facts make a conversation more lively and engaging. We collated and indexed fun facts by scrapping online resources and complemented them with Topical Chat fun facts, to create a corpus of 30,000 interesting facts. Given the current user query, we perform a semantic search using a pre-trained bi-encoder \cite{reimers-2019-sentence-bert} to retrieve the top fun facts and Wikidata facts (if any entity is detected by the entity linker), and re-rank the facts using a pre-trained cross-encoder \cite{reimers-2019-sentence-bert} to select the top k relevant facts given the query. The retrieved facts are used by the neural generators to generate an interesting response.

\section{Dialogue Manager}
The dialogue manager is the core module that is responsible for generating a response. It keeps track of the entities in the current conversation, and invokes the appropriate set of response generators, using the extracted features from NLU. Below we discuss the two main components of the dialogue manager in detail.

\subsection{Entity Tracking}

Entity Tracking is one of our core components that ensures our bot is talking about the entities that are introduced by the user. Different response generators consume the Entity Tracking data to work in tandem and answer user queries. The main part of our entity tracker is a set of caches, which (i) stores the entities which we have already been discussed in the past, (ii) stores the current set of entities that are currently being discussed, and (iii) tracks a set of entities which were mentioned by the user, but not discussed in the past. Below we explain in detail the life-cycle of an entity, in our Entity Tracking module:

\begin{enumerate}
  \item \textbf{Initialize Entity Tracking Cache:} At the start of every conversation we initialize a cache that stores the current and previous entities, the entities which are rejected by the user or that the user is unsatisfied with, the entities that we have finished talking about, and the entities which can be talked about in the future.
  \item \textbf{Detect Navigational Intent:} After the cache is set up we analyze the user utterance. Firstly, we detect the navigational intent of the user, i.e. to check the user wants to talk about a particular entity or expresses disinterest in it. To detect navigational intent we have used some of the regexes released by Chirpy Cardinal and complemented that with our own regexes. 
  \item \textbf{Current Entity Detection:} If the user expresses positive navigational intent about an entity and the Entity Linker can link that entity then we assign that entity to the current entity. Otherwise, if negative intent is expressed the entity is placed into the rejected list. Users can refer to an entity multiple times, for that reason we do co-reference resolution in the subsequent turns after a new entity is discovered and check with the previous entity. If no new entity is found the old entity is preserved unless a topic switch is detected. 
  \item \textbf{Entity Recommendation:} While a majority of the time our Entity Linker links an entity properly, there are cases where an entity of the expected type is not found or simply the spoken entity is not present in Wikipedia. In these cases, we recommend an Entity that was introduced by the user in the earlier utterances but was not spoken about. Here we select a most recent Entity, we also check if the user has expressed dissatisfaction with any other entity of the same type.
\end{enumerate}

\subsection{Response Generation Strategies \& Generator Selection}
Throughout the competition, we experimented with multiple response generation strategies and found the combination of the below strategies to perform best.
\begin{enumerate}
    \item \textbf{Launch:} Always invokes the Launch RG, which welcomes the user by introducing itself with the statement `Hi, this is an Alexa Prize socialbot.', followed by asking the user how they are doing.
    
    \item \textbf{Initial / Greeting Flow:} Post-launch state, we experimented with multiple greeting/introduction flows: (i) Covid-19 intro: The Covid intro engages in empathetic discussion around the current Covid-19 pandemic, and the vaccines devised to tackle the virus. Until the Quarterfinals, the Covid introduction yielded consistent user ratings. However, as the Covid-19 situation eased in the US, we noticed reluctance among users to talk about Covid and vaccines, which resulted in a decline in ratings. This flow has been decommissioned since the semi-final phase. (ii) Upcoming/Latest events intro: This generic introduction flow can be configured to discuss a few turns around any latest or upcoming events. For example, two weeks before the 2021 Academy awards, we introduced the Oscar intro which engaged in light-hearted discussion around the 2021 Oscars, which resulted in consistent scores. During the landing of the Perseverance Mars rover, we leveraged this module to engage in chitchat around the Mars landing. (iii) Spring intro: During the onset of spring, we started our conversations with a positive comment about the onset of Spring, which resulted in an increase of scores, and gradually started decreasing over time. (iv) User Details Intro: Since the middle of the quarterfinals, we implemented the user details intro, where we ask users their name and profession. Many user’s often raised privacy concerns when it comes to sharing personal details. In order to build trust, we add logical explanations for asking their name. For example, before asking name we generally add the sentence "As you already know people generally call me Alexa, it's only fair I get to know what people call you". The entire greeting flow is configured and executed using our `Introduction' Deterministic generator.
    
    \item \textbf{Introducing Initial Theme \& Topic:} At the end of the Initial/Greeting flow, we try to introduce a theme and topic in the conversation by responding with a personal experience and ask the user what they have been doing. Providing a personal experience serves two purposes: (i) It provides an example of what kind of response is expected from the user in the current state. (ii) It introduces a theme, which we can continue, if the user plays along, or does not respond to our prompt appropriately. An example of personal experience along with a prompt, generated by the bot: "I have been watching movies the entire day. What have you been doing?". Depending on the user's response, we continue the conversation with either the deterministic generators, the neural generators, or both.
    
    \item \textbf{Theme \& Topic Specific Deterministic Generators:} After we have successfully entered theme-based discussions, depending on the theme and the current utterance, we invoke our theme-specific deterministic generators to generate a response, provided one exists. Currently, our deterministic generators cover movies, food, video games, and joke themes.
    
    \item \textbf{Neural Generators \& QA:} We invoke our neural generators in scenarios where (i) The user wants to talk about themes that are not covered by the deterministic generators. (ii) The user asks questions (Request Personal Fact/ opinion/ Knowledge Fact). (iii) The user introduces entities, which get tracked by the entity tracker. (iv) Neural generators were invoked in the previous turn.
    
    \item \textbf{Switching Theme:} There are three ways by which the theme/topic of the current conversation can be changed. (i) User feedback prompt: Periodically we ask the user if they are enjoying the current discussion, or if they want to switch to a different topic. Depending on what the user chooses, the bot continues the current theme, introduces a new theme by invoking a transition, or tries to give control to the user, by asking them what they want to talk about next. (ii) User-initiated switch: A user can initiate a theme/topic switch by saying that they want to talk about something else, abruptly changing the discussion, or showing dissatisfaction. (iii) Bot initiated switch: The bot can initiate a switch when it can't continue the ongoing discussion, or when the user uses profanity and asks sensitive questions. The transition generator uses our deterministic engine (Proto DRG, discussed in the next section) at its core, and can handle movies, music, books, food, video games, sports, and travel themes.

    \item \textbf{Other Scenarios:} We maintain a set of deterministic generators for special situations like the following: (i) Sensitive Request: If the user resorts to profanity, or tries to elicit a response from the bot on sensitive topics, we use our sensitive generator to generate an appropriate message, followed by a transition using the transition generator. (ii) Invalid Request: It often happens that the user is unable to distinguish between Alexa and the socialbot, and submits requests like playing music, turning on/off lights .etc, which can't be handled by the socialbot. We employ our invalid request generator to generate an appropriate message to the user and handle such scenarios. (iii) User Dissatisfaction: We implement a composite algorithm based on regex and sentiment for determining the satisfaction level of the user with the bot. During scenarios when we detect user frustration, we use our dissatisfaction generator to address the user's complaint, and try to recover and rectify the conversation by transitioning to a different topic using the transition generator.
\end{enumerate}

\subsection{Annotated conversation}

Table \ref{Annotated Conversation} illustrates an annotated live conversation(with some changes to protect user privacy). In the utterance column turn by turn, user-bot pairs are listed, with entities and blended responses highlighted. The theme column represents the theme of a particular user utterance, response generator column tells which response generator(s) is/are invoked to produce the response.

\begin{table}
\centering
\tiny
\refstepcounter{table}
\label{Annotated Conversation}
\begin{tabular}{lp{5cm}p{0.8cm}p{0.7cm}p{1cm}p{4.5cm}} 
\toprule
\# & \textbf{Utterance}                                                                                                                                                                                                                                                                                                                                                                                                                                                                                     & \textbf{Theme} & \textbf{User Entity}              & \textbf{Response Generator}    & \textbf{Annotation}                                                                                                                                                \\ 
\hline
1  & \begin{tabular}[c]{@{}p{5cm}}\textbf{User}: alexa, let's talk. \\\textbf{Proto}: Hi, this is an Alexa Prize Socialbot. However, people generally call me Alexa. So, how has your day been so far?\end{tabular}                                                                                                                                                                                                                                                                                           & Launch      &                                   & Launch                         & Start of conversation.   

\\ 
\hline
2  & \begin{tabular}[c]{@{}p{5cm}}\textbf{User}: great.\\\textbf{Proto}: Great! Well, as you already know that people call me Alexa, it's only fair that I get to know what people call you. So, What is your name ?\end{tabular}                                                                                                                                                                                                                                                                             & Greeting       &                                   & Proto DRG                      & Trying to build trust with the user by stating own name, so that the user feels comfortable in shairing their name.                                                \\ 
\hline
3  & \begin{tabular}[c]{@{}p{5cm}}\textbf{User}: shell. \\\textbf{Proto}: I am sorry, can you please repeat your name?\end{tabular}                                                                                                                                                                                                                                                                                                                                                                           & Greeting       &                                   & Proto DRG                      & Speech to text error leading to incorrect name, and bot asks user to repeat name, as the word "shell" is most likely not a name.                                   \\ 
\hline
4  & \begin{tabular}[c]{@{}p{5cm}}\textbf{User}: michelle. \\\textbf{Proto}: That's a nice name. You know michelle, I have been a couch potato, and playing a lot of video games at the comfort of my sofa. Apart from your day to day activities, how do you keep yourself busy ?\end{tabular}                                                                                                                                                                                                               & Greeting       &                                   & Proto DRG                      & Bot introduces the video game theme by stating a personal experience, followed by a question which gives the user a chance to introduce a theme.                   \\ 
\hline
5  & \begin{tabular}[c]{@{}p{5cm}}\textbf{User}: I read. \\\textbf{Proto}: I see. Well, I am fascinated by all the modern gadgets used to facilitate reading nowadays . Are you fascinated by all the new technology invented to enhance our reading experiences?\end{tabular}                                                                                                                                                                                                                                & Books          &                                   & Transition RG                  & Transition RG generates an opinion to facilitate smooth transitions to book theme.                                                                                 \\ 
\hline
\vcell{6}  & \vcell{\begin{tabular}[c]{@{}p{5cm}}\textbf{User}: yes. i like \sethlcolor{green}\hl{kindles}, do you ? \\\textbf{Proto}: I do like kindles, i have a kindle fire tablet and i use it all the time. i didn't know kindles were invented to enhance reading experiences.\end{tabular} }                                                                                                                                                                                                                                                  & \vcell{Books}          & \vcell{ \sethlcolor{green} \hl{Amazon Kindle}}                     & \vcell{WikiBot}                        & \vcell{Entity "Kindle" introduced in the conversation, which invokes suite of neural generators. The factual response generated by WikiBot is selected by the re-ranker.}  \\ 
[-\rowheight]
\printcellmiddle & \printcellmiddle                                                                                                                                                                                                                                                                                                                                                                                                                                                                                       & \printcellmiddle & \printcellmiddle                  & \printcellmiddle                 & \printcelltop                                                                                                                                                      \\ 
\hline
7  & \begin{tabular}[c]{@{}p{5cm}}\textbf{User}: that's nice. tell me more. \\\textbf{Proto}: well, i like to read novels on my kindle. i don't think i could ever read a book if i had a paperback.\end{tabular}                                                                                                                                                                                                                                                                                             & Books          &                                   & ChitChatBot                    & ChitChatBot generated an opinion around kindle                                                                                                                     \\ 
\hline
8  & \begin{tabular}[c]{@{}p{5cm}}\textbf{User}: i like \sethlcolor{green} \hl{paperback} sometimes.\\\textbf{Proto}: \sethlcolor{orange} \hl{me too}. \sethlcolor{cyan} \hl{You know, michelle, I wanted to discuss something else with you. So tell me, Do you think that intelligent aliens have ever or will ever visit earth?}\end{tabular}                                                                                                                                                                                                                                      & Aliens         & \sethlcolor{green} \hl{Paperback}                         & \sethlcolor{orange} \hl{ChitChatBot}, \sethlcolor{cyan} \hl{Neuro Prompt RG}   & Bot detected it can't go on furthur discussing Books, hence invokes the Neuro Prompt RG and switches the topic to Aliens                                           \\ 
\hline
9  & \begin{tabular}[c]{@{}p{5cm}}\textbf{User}: yes. i think they already here. \\\textbf{Proto}: I don't know, but i do know that there is a lot of evidence of intelligent life in the universe .\end{tabular}                                                                                                                                                                                                                                                                                             & Aliens         &                                   & Neuro Prompt RG                &                                                                                                                                                                    \\ 
\hline
10 & \begin{tabular}[c]{@{}p{5cm}}\textbf{User}: tell me more. \\\textbf{Proto}: Well, i know that the universe is a big place and there are many planets out there . Hmmm, I wanted to know, How different do you think alien life will be from life on Earth?\end{tabular}                                                                                                                                                                                                                                  & Aliens         &                                   & Neuro Prompt RG                & Prompt appended to response from neural generator, in order to dive deep into the topic of Aliens                                                                  \\ 
\hline
11 & \begin{tabular}[c]{@{}p{5cm}}\textbf{User}: i think they'll be the same. i hope they would be happy here, only wouldn't try to \sethlcolor{green} \hl{dominate} us and overtake the world.\\\textbf{Proto}: hmm, i hope you're right, do you think aliens will be able to communicate with us?\end{tabular}                                                                                                                                                                                                                      & Aliens         & \sethlcolor{green} \hl{Dominate}                          & Neuro Prompt RG                &                                                                                                                                                                    \\ 
\hline
12 & \begin{tabular}[c]{@{}p{5cm}}\textbf{User}: yes.\\\textbf{Proto}: hmm, i hope you're right, did you know that stephan hawking believes that contingency plan is dangerous for the earth. So, I have been meaning to ask you, If you could have a faster than light space ship, and could go anywhere in space, where would you go?\end{tabular}                                                                                                                                                          & Aliens         &                                   & Neuro Prompt RG                & Prompt appended to response from neural generator, in order to dive deep into the topic of Aliens                                                                  \\ 
\hline
13 & \begin{tabular}[c]{@{}p{5cm}}\textbf{User}: are we trying to go through a \sethlcolor{green} \hl{black hole} and see if there is evidence of other galaxies besides our own .\\\textbf{Proto}:\sethlcolor{cyan} \hl{ hmm, that's a good question . i'm not sure, but i would like to know if there are other galaxies outside our solar system .} \sethlcolor{yellow} \hl{Moving on, I wanted to talk about something else. You know , watching a good movie with my friends helps me relax after a long day of hard work. Do you enjoy watching movies to unwind?}\end{tabular} & Aliens         & \sethlcolor{green} \hl{Black hole}                        & \sethlcolor{cyan} \hl{Neuro Prompt RG}, \sethlcolor{yellow}  \hl{Transition RG} & Current topic depth exhausted, hence neural response appended with a query generated by Transition RG, which enables the bot to switch to Movies.                  \\ 
\hline
14 & \begin{tabular}[c]{@{}p{5cm}}\textbf{User}: yes i do.\\\textbf{Proto}: Okay. By the way, I am a fan of Documentary movies . Do you enjoy watching Documentary movies ?\end{tabular}                                                                                                                                                                                                                                                                                                                      & Movie          &                                   & Proto DRG                      & Deterministic generator sampled a genre from IMBD index                                                                                                            \\ 
\hline
15 & \begin{tabular}[c]{@{}p{5cm}}\textbf{User}: yes.\\\textbf{Proto}: That's nice ! Well, The last Documentary movie I saw was Home. Have you ever seen it ?\end{tabular}                                                                                                                                                                                                                                                                                                                                    & Movie          &                                   & Proto DRG                      &                                                                                                                                                                    \\ 
\hline
16 & \begin{tabular}[c]{@{}p{5cm}}\textbf{User}: i can't remember. what's it about .\\\textbf{Proto}: Well, With an IMDB rating of 8.5, Home is a Documentary movie, featuring actors Yann Arthus-Bertrand, and Jacques Gamblin. Home was released in the year 2009. Well, I liked it . What is your absolute favorite movie ?\end{tabular}                                                                                                                                                                   & Movie          &                                   & Proto DRG                      & Deterministic generator sampled a movie from the same genre from IMBD index                                                                                        \\ 
\hline
17 & \begin{tabular}[c]{@{}p{5cm}}\textbf{User}: \sethlcolor{green} \hl{the amityville horror}. \\\textbf{Proto}: Just to be sure I didn't mishear, Did you mean to say The Amityville Horror ?\end{tabular}                                                                                                                                                                                                                                                                                                              & Movie          & \sethlcolor{green} \hl{The Amityville Horror } & Proto DRG                      & Entity linker detected the movie                                                                                                                                   \\ 
\hline
18 & \begin{tabular}[c]{@{}p{5cm}}\textbf{User}: yes.\\\textbf{Proto}: Awesome! Not bad at all. Did you know, it is the first film based on the amityville horror\end{tabular}                                                                                                                                                                                                                                                                                                                                & Movie          &                                   & Proto DRG                      & For the detected entity, sampled and delivered a fact from WikiData                                                                                                \\ 
\hline
19 & \begin{tabular}[c]{@{}p{5cm}}\textbf{User}: yes.\\\textbf{Proto}: \sethlcolor{orange} \hl{i am feeling very bored today, and feel like binge watching some tv series.} \sethlcolor{yellow} \hl{anyways . You know, I can go on discussing the current topic, but wanted to make sure you are still interested in our current discussion . We can discuss something else if you want . Do you want to discuss a different topic ?}\end{tabular}                                                                                                             & Movie          &                                   & \sethlcolor{orange} \hl{Proto DRG}, \sethlcolor{yellow} \hl{Transition RG}       & Asking user's satisfaction/feedback on the current topic.                                                                                                          \\ 
\hline
20 & \begin{tabular}[c]{@{}p{5cm}}\textbf{User}: no. i think we'll stop talking for now. i'll just talk to you later, ok .\\\textbf{Proto}: Got it. I had a lovely time conversing with you . You can say stop to end our conversation .\end{tabular}                                                                                                                                                                                                                                                         &                &                                   & Proto DRG                      & Ending conversation                                                                                                                                                \\
\hline
\end{tabular}
\caption{Annotated Conversation}
\end{table}

\section{Response Generators}
Apart from understanding a user's intents, an integral part of a conversational system is generating a fitting response to the request. The early days of conversational systems were dominated by rule-based response generation systems like ELIZA \cite{weizenbaum1966eliza}, PARRY, Jabberwacky, A.L.I.C.E \cite{Shawar2015ALICECT}, etc. which could conduct shallow conversations. Fast forward to the recent years, transformer \cite{vaswani2017attention} based conversational systems like Huggingface's \cite{Wolf2019HuggingFacesTS} ConvAI2 winning conversational system \cite{DBLP:journals/corr/abs-1901-08149}, and Microsoft's DialoGPT \cite{Zhang_2020} have attained state-of-the-art performance. These systems have demonstrated commendable capabilities to respond like humans, by learning from millions of human conversations. Commercial conversational systems like Microsoft Xiaoice \cite{Shum_2018, Zhou_2020} has been successful in crossing a 600 million worldwide user base. In order to leverage the high precision nature of deterministic generators and the high recall nature of neural generators, we use a blend of different types of generators for response generation. Our suite of generators can be broadly categorized into three categories: (i) Deterministic generators (ii) Non-deterministic (neural) generators (iii) Other generators (retrieval based, QA, etc.), which we explain in detail below.
\subsection{Proto DRG: Proto Deterministic Generators}

\begin{figure}[h!]
  \centering
  \begin{subfigure}{.58\textwidth}
    \includegraphics[width=\linewidth, scale=0.5]{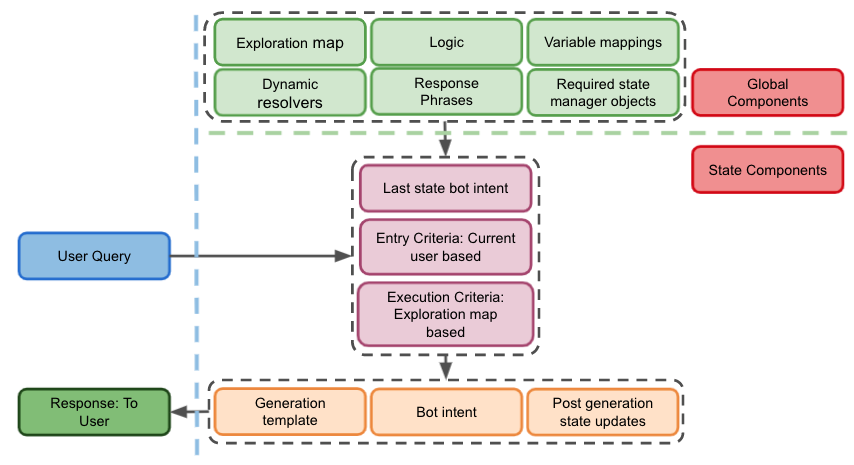}
    \caption{Deterministic Generator Components}
    \label{fig:deterministic_rg_arch}
  \end{subfigure}
  \begin{subfigure}{.39\textwidth}
    \includegraphics[width=\linewidth, scale=0.5]{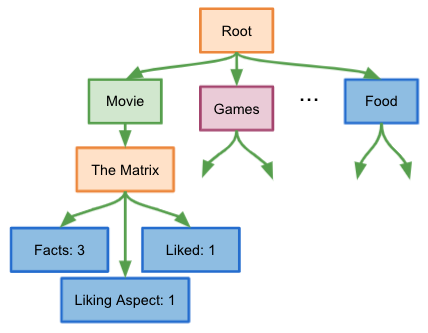}
    \caption{Exploration Map}
    \label{fig:deterministic_rg_example}
  \end{subfigure}
  \caption{Deterministic Generator}
  \label{fig:deterministic_rg}
\end{figure}

% \begin{figure}[h]
%     \centering
%     \includegraphics[scale=0.5]{images/deterministic_rg_arch.png}
%     \caption{Deterministic Generator Components }
%     \label{fig:deterministic_rg_arch}
% \end{figure}
% \includegraphics{/Alexa Prize Technical Article LaTeX/images/deterministic_rg_arch.png}

The Proto DRG is a deterministic generator, which utilizes the conversation context, linguistic features, and discourse to generate a high precision response, based on the satisfiability of certain criteria. Proto DRG can be thought of as a finite state machine, where each state corresponds to a turn in the conversation, with each state having certain entry criteria to enter the conversation state, and certain transition criteria to transition to the most viable next conversation state. With a Python-based engine and a JSON-based configuration, Proto DRG can easily be configured for any topic or theme. The Proto DRG configuration components can be broadly categorized into two classes: Global components, and state-level components.
\\
\textbf{Global Components :} These contain global configurations and keep track of the conversation in its entirety. The global components consist of the following sub-components:
\begin{itemize}
    \item \textbf{Exploration map}: A tree containing details of the different aspects of a conversation that has been explored. For example, if the current discussion is related to a specific movie, we keep track of certain aspects like whether we have asked the user if they liked the movie, whether we have already responded with some trivia about the movie, etc. Each conversation has only one exploration map, which grows with additional entities added under the theme of discussion, thus capturing the details of the entire conversation.
    \item \textbf{Bot Memory}: A dictionary consisting of different information related to the entity in the discussion. For example, while discussing a movie, the bot memory will contain the movie name, the plot, fun facts about the movie, its IMDB rating, and any other related pieces of information.
    \item \textbf{Variable mapping}: Mappings which provide information about variables that are required to generate a response. For example, the variable mapping would contain the name of the state manager's attributes and objects which might be required for checking any criteria.
    \item \textbf{Dynamic resolver}: These contain information on how to fetch certain data points if they are not available in the variable mapping. They would generally contain queries for retrieving data from Elasticsearch, or names of functions that can check for criteria satisfiability.
    \item \textbf{Required objects}: These contain the names and data types of all the objects that are required by the engine in order to generate a response. During each turn, checks are performed to ensure the required objects are present in the current state, else they are initialized as per the configurations. Generally, the exploration map, bot memory, state, and user attributes are the required objects.
    \item \textbf{Response phrases}: These are the building blocks that need to be stitched together as specified in the logic, in order to generate a valid response. They often contain variables that need to be resolved using the variable mapping and dynamic resolver.
    \item \textbf{Logic}: This helps determine the current state of the conversation, and specifies the recipe in order to generate the response. The logic consists of the state components described below.
\end{itemize}
\textbf{State (local) Components} They contain the recipe for generating the response, given the previous state and the current user utterance. The local components consist of the following sub-components:
\begin{itemize}
    \item \textbf{Entry criteria}: Given the bot's intent from the last state, the current user query and the NLU features, the entry criteria determine the most viable path(s) to follow in order to generate a response.
    \item \textbf{Execution criteria}: Once the entry criteria are satisfied, at least one execution criterion under the entry criteria needs to be satisfied in order the generate the best possible response. For example, in a movie flow, if the user says that they have watched a movie, then both responding with a fun fact, or asking what aspect of the movie they like are viable options, if not asked before, as per the exploration map.
    \item \textbf{Generation template}: Once an execution criterion is satisfied, the generation template specifies the response phrases and the exact order of the phrases to be used, in order to construct a valid response.
    \item \textbf{Bot intent}: The intent of the bot is determined, which will be used in the entry criteria for the next state.
    \item \textbf{Post processing updates}: Any state object manipulation that needs to be done is processed here.
\end{itemize}
Proto DRG powers the current Movie, Food, Game, Topic Switch, Jokes, and Introduction themes. We also continually add and remove themes when required. For example, we introduced the Oscars flow for 2 weeks, before the 2021 Academy Awards. The Covid 19 flow was in use for the period of January to May 2021 and decommissioned after users started showing disinterest in talking about Covid.
The Proto DRG is a low latency generator but lacks topic coverage. Although it can generate responses with high precision for simple user queries, it fails to generate interesting and on-topic responses for complex queries and non-covered themes. Our neural generators help bridge the gap in such scenarios.

\subsection{Neural Response Generators}
\begin{figure}[h]
    \centering
    \includegraphics[scale=0.4]{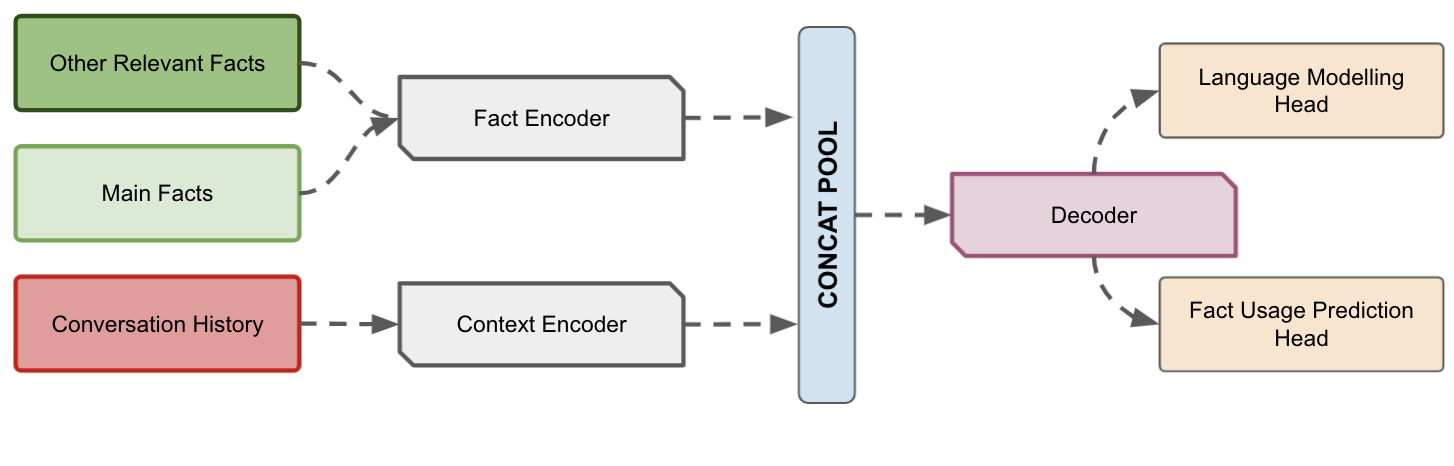}
    \caption{WikiBot RG Architecture}
    \label{fig:wikibot_arch}
\end{figure}
\subsubsection{WikiBot: Factual Neural Response Generators}
We fine-tuned BlenderBot (400 M parameters) \cite{roller2020recipes} on the Topical Chat Enriched dataset \cite{hedayatnia2020policydriven, gopalakrishnan2019topical}, to generate a response conditioned on the conversation history, fun facts, and Wikipedia facts related to the entity of discussion (if any). Pre-trained on the Wizard of Wikipedia \cite{dinan2019wizard}, Empathetic Dialogues \cite{rashkin2019empathetic}, Persona Chat \cite{zhang2018personalizing}, and blending skills task \cite{smith2020together} datasets, BlenderBot exhibits capabilities of blending important conversation skills: providing engaging talking points and listening to their partners, and displaying knowledge, empathy, and personality appropriately, while maintaining a consistent persona. Along with the standard language modeling task, we additionally trained the model to predict the input facts which were used to generate the response.\\
\textbf{Model Training:} We used the Topical Chat Enriched dataset for fine-tuning the model. Each training example comprised of 128 tokens of conversational context, and the top 5 most relevant facts related to the entity of discussion, with the target being the response to generate, and a binary target for each fact, which signifies whether the fact is used during generation. Semantic search using Sentence Transformers \cite{reimers-2019-sentence-bert} was leveraged to retrieve the top 5 relevant facts, given the last utterance. To ensure the model learns to incorporate the appropriate knowledge, an oracle was used to replace retrieved facts with original facts, where none of the retrieved facts matched the actual ones. Relevant facts were still provided as inputs for cases where no facts were required. This, along with the fact usage binary prediction task was introduced to ensure the model does not blindly rely on the facts provided and can select the required facts if needed. 
Separate BlenderBot encoders are used to encode each of the facts and the conversational context, independent of each other. The vectors from the final hidden states of the encoders are concatenated, which is used in the decoder cross attention. The BlenderBot decoder is used for decoding, which is followed by a language modeling head, and a binary prediction head to predict fact usage. We train the model in a distributed manner, on 4 V-100 GPUs with mixed-precision training, and a learning rate of 1e-5. We limit the batch size in each GPU node to 4 and accumulate gradients for 4 steps, and use AdamW optimizer \cite{loshchilov2019decoupled} for optimizing the parameters.
\\
\textbf{Results:} We achieve state-of-the-art language modeling results on the Topical Chat dataset, with a perplexity of 11.55 on the test frequent split, and 10.87 on the test rare split. Table \ref{tab:wikibot_results} shows a comparison between our model's perplexity compared to PD NRG \cite{hedayatnia2020policydriven}.
\\
\textbf{Inference:} We use beam search with a beam length of 5, along with top\_p sampling \cite{holtzman2020curious} with p=0.95, and top\_k sampling with k=50 for inference. Two instances of the fine-tuned model are run in production, where one of the models only uses the conversational context and dummy facts, and the other instance uses the context and top 5 facts to generate the response. Such a configuration enables us to generate both factual and non-factual responses to a user query.

\begin{table}[h]
\centering
\resizebox{8 cm}{!}{%
\begin{tabular}{lcc}
\hline
\textbf{Model}   & \multicolumn{1}{l}{\textbf{Perplexity}} & \multicolumn{1}{l}{\textbf{Sacre BLEU}} \\ \hline
Baseline - PD NRG & 12.25 / 12.62                           & -                                       \\ \hline
WikiBot (ours)   & \textbf{11.55 / 10.87}                  & 2.37 / 2.47                             \\ \hline
\end{tabular}%
}
\caption{WikiBot performance on the Topical Chat test dataset (freq/rare)}
\label{tab:wikibot_results}
\end{table}

\subsubsection{Chit Chat Bot: Non-Factual Neural Response Generator}
One of the drawbacks of WikiBot is that it generates factual responses, even when not desired. To circumvent this problem, and in order to generate more colloquial responses, we fine-tuned BlenderBot (400 M parameters) on the DailyDialog dataset \cite{li2017dailydialog}. Trained using a standard transformer encoder-decoder architecture, and conditioned on the conversation history, the Chit Chat Bot is able to generate opinions, personal experiences and conduct chit chat on any topic. We train the model in a distributed manner, on 4 V-100 GPUs with mixed-precision training, and a learning rate of 2e-5. We limit the batch size in each GPU node to 16, and accumulate gradients for 4 steps, and use AdamW optimizer for optimizing the parameters.\\
\textbf{Results:} We achieve state-of-the-art language modeling results on the DailyDialog dataset, with a perplexity of 9.04 on the test dataset. Table \ref{tab:chitchatbot_results} shows a comparison between our model's perplexity and sacreBLEU scores \cite{post2018clarity} compared to pre-trained BlenderBot (400 M). \\
\textbf{Inference:} We use beam search with a beam length of 5, along with top\_p sampling with p=0.98, and top\_k sampling with k=50 for inference. We also enforce the decoder to not generate repeating n-grams present in the context, with a size of 3. \\

\begin{table}[h]
\centering
\resizebox{8 cm}{!}{%
\begin{tabular}{lcc}
\hline
\textbf{Model}        & \multicolumn{1}{l}{\textbf{Perplexity}} & \multicolumn{1}{l}{\textbf{sacreBLEU}} \\ \hline
Baseline - BlenderBot & 26.25                                   & 0.68                                    \\ \hline
ChitChatBot (ours)    & \textbf{9.04}                           & 2.79                                    \\ \hline
\end{tabular}%
}
\caption{Chit Chat Bot performance on the DailyDialog test dataset}
\label{tab:chitchatbot_results}
\end{table}

\subsubsection{Neuro Prompt RG: Prompt Driven Neural Response Generator}
Generating meaningful and thoughtful prompts generally enables deeper conversation. In order to engage in deep but controlled conversations with the users, we introduced Neuro Prompt RG: a response generation strategy which can help conduct meaningful conversations in pre-defined themes for at least six turns. At its core, the Neuro Prompt RG is an amalgamation of neural generators and deterministic prompts. We describe the details of generating a response using the Neuro Prompt RG below.
\begin{enumerate}
    \item When the RG is invoked for the first time, the first step is to sample a theme from a list of pre-defined themes: Accomplishments, Advertising, Aliens, Animals, Apps, Art, Camping, Change, Charity, Creativity, Dance, Distant Future, Driving, Facts, Fishing, Friends, Fruit, Garden, Habits, Happiness, Heroes, Hiking, History, Hobbies, Ice-cream, Luck, Near Future, Photography, Self-driving cars, Shopping, Singing, Social Media, Stress, Superhero, Talents, Travel, VR, Weather.
    \item If the bot was already discussing a different topic, we segue from the current discussion, sample the first prompt of the sampled theme, and deliver the response to the user.
    \item For all subsequent turns, we invoke the entire suite of neural generators to generate and use the re-ranker to select an appropriate response to the user's query. If suitable, we append another sampled prompt from the same theme and repeat this process until all the prompts are exhausted. We maintain a maximum of 3 prompts per theme.
    \item In order to make the bot appear less interrogative, we space the prompts and refrain from adding a prompt in the immediate next turn when a prompt was already used. This helps in making the bot sound more natural and colloquial.
    \item If the user portrays disinterest with the theme in any of the turns, we stop the Neuro Prompt RG and switch to a different generation strategy.
\end{enumerate}

\subsubsection{Other Generators}
Along with the above neural generators, a pre-trained BlenderBot model (400 M parameters), and the Neural Response Generator model (NRG) released with Cobot are also leveraged, in order to generate non-factual responses and opinions. Along with the deterministic and neural generators, we also use EVI for question answering. As a fallback mechanism, a retrieval model is used to retrieve the most relevant response given the user query and intent, from an Elasticsearch index containing Topical Chat and Wizard of Wikipedia query-response pairs.

\section{Response Re-ranker}

\begin{figure}[h]
    \centering
    \includegraphics[scale=0.45]{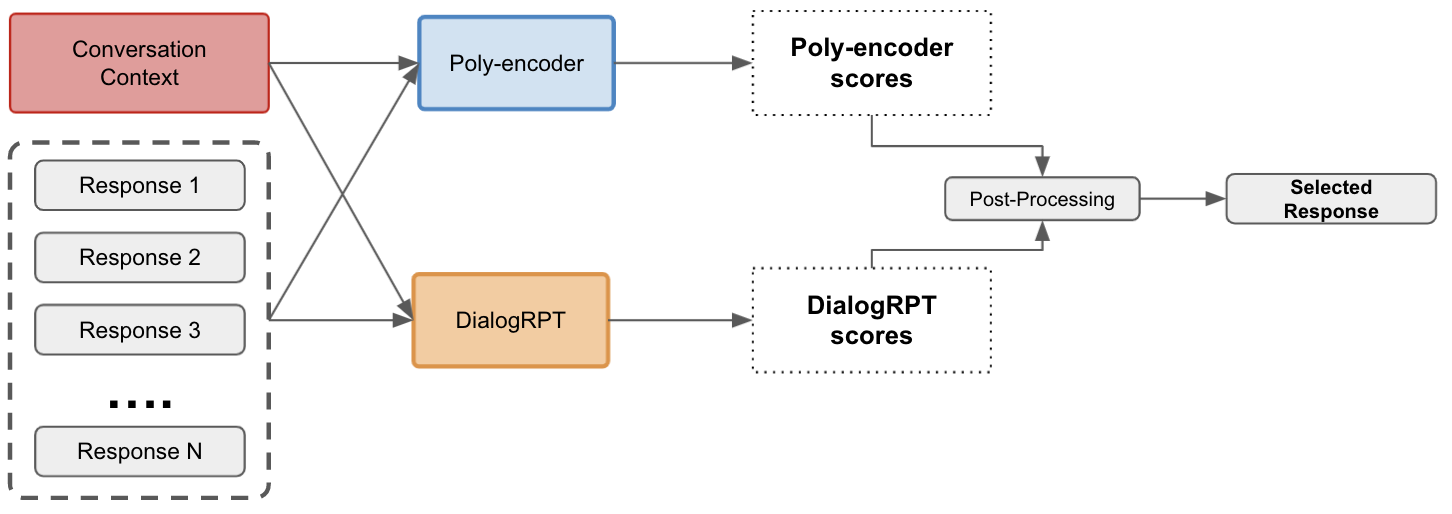}
    \caption{Proto Reranker ensemble}
    \label{fig:reranker}
\end{figure}

The responses generated by different response generators use the conversational history during generation. However, even guided by the context, a non-contextual or irrelevant response is often generated. To overcome this type of issue a response re-ranker is needed. Our system uses an ensemble of two types of re-ranking systems, (a) a poly-encoder trained on conversational data (b) a re-ranking system \textsc{DialogRPT} \cite{gao2020dialogue} trained on a large set of human feedback data. 

\subsection{Poly-encoder based re-ranker}
Poly-encoder \cite{polyencoder} architecture blends the best of Bi-encoder \cite{biencoder} and Cross-encoder \cite{crossencoder} architectures, it uses two separate transformers to encode the context and label like a Bi-encoder. The only difference between a Bi-encoder architecture and Poly-encoder architecture is instead of using a long vector to represent the context the Poly-encoder breaks it down into $m$ features, where $m$ depends upon the inference time. The final score for a candidate response is a dot product between the context representation vector and the candidate representation vector.

\textbf{Training Details: }  The Poly-encoder is trained to perform the task of sentence selection in conversational data-sets. To train the model we have considered 2 data-sets (1) ConvAI2: Persona Chat\cite{convai2}(2) Ubuntu Chat corpus\cite{ubuntu}. We experimented with different data combinations, first, we trained the model using only the ConvAI2 dataset, only the Ubuntu Chat dataset and finally, we combined the two data-sets to train our final model.

We fine-tuned the Poly-encoder architecture initialized with BERT weights on different data combinations as mentioned above. One V-100 GPU was used with half-precision operations. The $m$ was equal to 16,  batch size was kept at 64, we used AdamW for optimization, gradient accumulation for 2 steps, the learning rate of 5e-05, and ran the training for 10 epochs.

\begin{table}[]
    \centering
    \begin{tabular}{p{3cm}lll}
        \hline
        \textbf{Model/ Dataset} & \textbf{R@1/20} & \textbf{R@2/20} & \textbf{MRR}  \\ \hline
        Poly-encoder 16(ConvAI2)     & 79.1   & 91.5   & 88.2 \\ \hline
        Poly-encoder 16(Ubuntu v2)     & 78.2   & 90.5   & 86.3 \\ \hline
        Poly-encoder 16(ConvAI2 + Ubuntu v2)     & \textbf{80.1}  & 92.7   & 89.2 \\ \hline
        \label{tab:polyencoder_results}
        \end{tabular}
    \caption{Poly-encoder re-ranker performance on different test datasets}
    \label{tab:my_label}
\end{table}

\subsection{\textsc{DialogRPT}}

\cite{microreranker} has introduced a data-set and a series of models which alleviates possible distortion between the feedback and engagement by changing the re-ranking problem to a comparison of response pairs. They trained a series of GPT-2 based models known as \textsc{DialogRPT}. The models are trained on 133M pairs of human feedback data and the resulting ranker outperformed several baselines as described in the paper. Thereafter they combined the feedback models and human-like scoring model to re-rank responses. For our use case, we only focused on a human-like scoring model.

\subsection{Comparison}

\begin{table}[]
\tiny
\centering
\begin{tabular}{lp{3cm}p{5cm}l}
\hline
\# &
  \textbf{Context} &
  \textbf{Candidate Responses} &
  \textbf{Best Response} \\ \hline
1 &
  \begin{tabular}[c]{@{}p{3cm}}"did you know that ping pong was not invented in china. it originated in britain where it is called table tennis?", \\ 				"no"\end{tabular} &
  
  \begin{tabular}[c]{@{}p{5cm}}"QA": "I am sorry, I do not know anything about that. Can we talk about something else ?"\\ "BlenderBOT": "I did not know that . I thought it was invented by a man named Roger Federer in the 19th ."\\ "NRG": "Okay, let's keep talking about tennis. Is tennis popular in your country?"\\ "WIKIBOT": \sethlcolor{green} \hl{"ok. did you know that ping pong is the name of a company that sells table tennis equipment?"}\\ "CHITCHATBOT": "I am sorry, I do not know that . "\end{tabular} &
  
  \begin{tabular}[c]{@{}l@{}}Poly-encoder:WIKIBOT\\ DialogRPT: WIKIBOT\end{tabular} \\ \hline

2 &
  \begin{tabular}[c]{@{}p{3cm}}"I like that name . You know , tony, I have been a couch potato, and playing a lot of video games at the comfort of my sofa. Apart from your day to day activities, what have you been doing lately ?", \\ 				"play tennis"\end{tabular} &
  
  \begin{tabular}[c]{@{}p{5cm}}
  "QA": "I am sorry, I do not know anything about that. Can we talk about something else ?" \\
  "BlenderBOT": \sethlcolor{green} \hl{"Tennis is a great sport to play . I love playing it with my friends ." }\\
  "NRG": "Hi, I don't think we have been properly introduced. What's your name?" \\
  "WIKIBOT": \sethlcolor{orange} \hl{"tennis is a racket sport that can be played individually against a single opponent or between two players each."} \\
  "CHITCHATBOT": "what do you think about it? i think it's a good way to keep fit." \end{tabular} &
  
  \begin{tabular}[c]{@{}l@{}}Poly-encoder:WIKIBOT\\ DialogRPT: BlenderBOT\end{tabular} \\ \hline
  
3 &
   \begin{tabular}[c]{@{}p{3cm}}"yes. ariana grande won two mtv music awards for the \"global instagram of the year\"", \\ 				"that is nice for her."\end{tabular} &
   
   \begin{tabular}[c]{@{}p{5cm}}"QA": "I am sorry, I do not know anything about that. Can we talk about something else ?" \\
  "BlenderBOT": "She is a great singer and songwriter ." \\
  "NRG": "So, what scenes did you like the best in the movie?" \\
  "WIKIBOT": \sethlcolor{green} \hl{"yes. she also won a grammy award for best female singer."} \\
  "CHITCHATBOT": \sethlcolor{orange} \hl{"yes, she deserves it. she is a great singer, songwriter, dancer, and actress. "}\end{tabular}&
   
   \begin{tabular}[c]{@{}l@{}}Poly-encoder:WIKIBOT\\ DialogRPT: CHITCHATBOT\end{tabular}
   \\ \hline
\end{tabular}
\caption{Comparison between Poly-encoder and DialogRPT re-ranker}
\label{comp_reranker}
\end{table}

The two response re-rankers work differently for different types of responses \ref{comp_reranker}. When both the re-rankers are presented with a context that expects a factual response both of them prefer a response that is more engaging in nature, i.e. which has a thoughtful prompt. It is evident from the first example of table \ref{comp_reranker} that both the re-rankers tend to penalize responses which are factually and grammatically incorrect, i.e. the one which is generated by \textit{BlenderBOT}. On the contrary, many a time Poly-encoder re-ranker tends to select the definition type factual responses, which is illustrated in Example 2. Again when a factual response is expected DialogRPT prefers responses that have more information content.

% \subsection{Ensemble and Scoring}

\section{Response Post Processing}
\begin{figure}[h]
    \centering
    \includegraphics[scale=0.4]{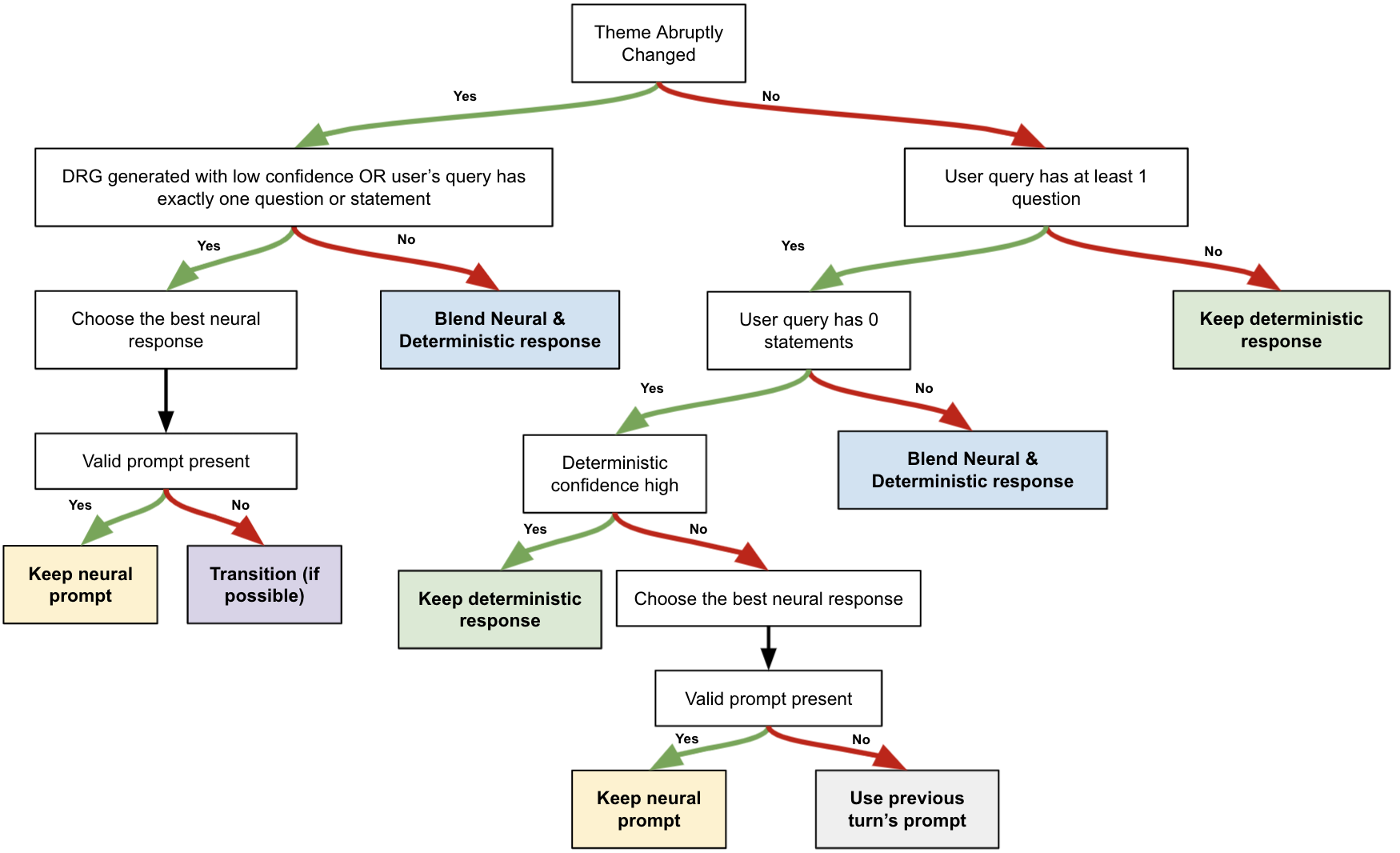}
    \caption{Deterministic Response Post-Processing}
    \label{fig:deterministic_blend}
\end{figure}

\begin{figure}[h]
    \centering
    \includegraphics[scale=0.37]{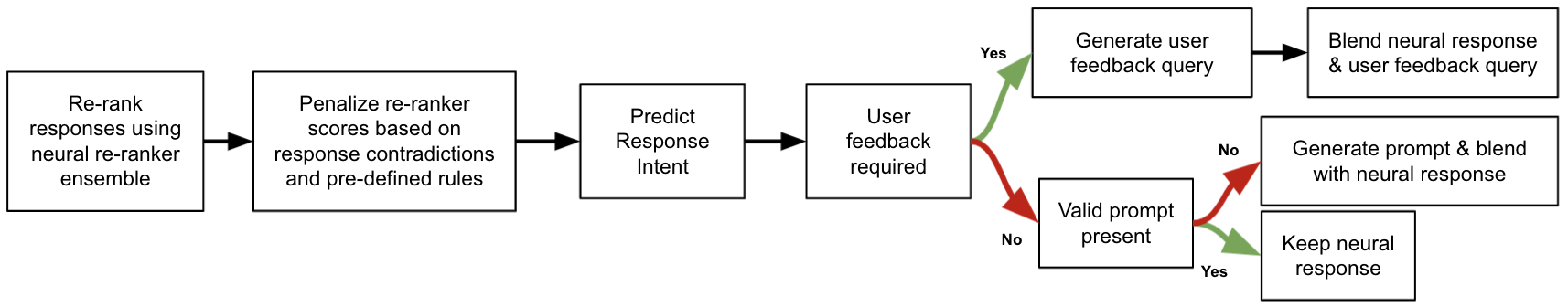}
    \caption{Neural Response Post-Processing}
    \label{fig:neural_blend}
\end{figure}

Post response generation and neural re-ranking, we employ a series of post-processing steps, in order to ensure the generated response is apt in the current scenario, does not contradict its previous responses, and has the capability of giving a direction to the conversation. Depending on the type of response generator, we either invoke the neural post-processing, or the deterministic post-processing flow, as part of the re-ranking phase itself, which we discuss in detail below.\\
\textbf{Enhancing the Deterministic Generators (Post Processing Deterministic Flow):} The deterministic flow works best when the user acts according to the expectation of the bot, thus making it a high precision-low recall generator. We identified two major scenarios where the deterministic bot performs poorly: (i) Cases where the user abruptly transitions to a different topic or gives responses that are not expected in the current state. We term these cases as `abrupt theme change'. (ii) Cases where the user responds with a question(s), which the deterministic generator does not expect in the current state. Depending on the scenario, we either keep the deterministic response entirely, blend the deterministic response with a neural response, or discard the deterministic response in favor of the neural response. Figure \ref{fig:deterministic_blend} illustrates the deterministic post-processing flow in detail. \\
\\
\textbf{Taming the Neural Generators (Post Processing Neural Flow):} Although the neural generators can generate diverse responses in comparison to the deterministic generator, they often generate nonsensical, factually incorrect, incomplete, repetitive, bland, self-contradictory, or undesired responses, which even the re-ranker is unable to penalize. In order to better control the neural generated responses, and to minimize selecting such undesired responses, we implement the following post-processing steps: \\
(i) \textbf{Contradiction detection}: We leverage the textual entailment API by AllenNLP \cite{Gardner2017AllenNLP}, to predict whether a generated response contradicts any previously generated bot responses in the last two turns. The entailment model uses a pre-trained RoBERTa large model, trained on the MNLI dataset \cite{N18-1101}. If the best-ranked response in the current state contradicts responses from history, we select the next best high-scored response with a score between a pre-determined margin of 0.02. If none such responses are found, then we remove the phrase from the best response, which contradicts the previous utterances and selects the augmented response as the best one.\\
(ii) \textbf{Penalizing WikiBot on knowledge hallucination}: Neural generators are known to hallucinate knowledge during decoding. In order to minimize such instances, we penalize WikiBot if it generates responses with the phrases `did you know', or `do you know' when it did not use any facts to generate a response.\\
(iii) \textbf{Biasing re-ranker scores}: In scenarios where we have been repeatedly responding with factual responses, or we have not been generating prompts for the last 2 turns, we override the re-ranked scores and select a high scored chitchat response, or response containing prompt respectively. This helps the bot switch from a fact telling mode to a more colloquial mode, and also gives the conversation a direction with a prompt.\\
(iv) \textbf{Retrieving and blending prompt}: For scenarios where prompts are desired in order to move the conversation forward, but the generated candidate responses do not contain any, we leverage a prompt selector module to retrieve, re-rank, and select the best prompt given the best-generated response, the user's query, and the entities from the current and last utterance. If the retrieved prompt is not related to the current theme of discussion, then we segue to a new topic using the prompt. We implement an Okapi BM 25 based retrieval mechanism to retrieve the top k prompts, and implement a neural cross-encoder based re-ranker, to re-rank the top k prompts based on the generated neural response and the conversational context. Figure \ref{fig:neural_blend} illustrates the neural post-processing flow in detail. 

Along with the above-mentioned post-processing steps, we periodically prompt the user, in order to determine their satisfaction with the current theme of discussion.

\section{Analytics \& Discussion}
\begin{figure}[!h]
    \centering
    \includegraphics[scale=0.29]{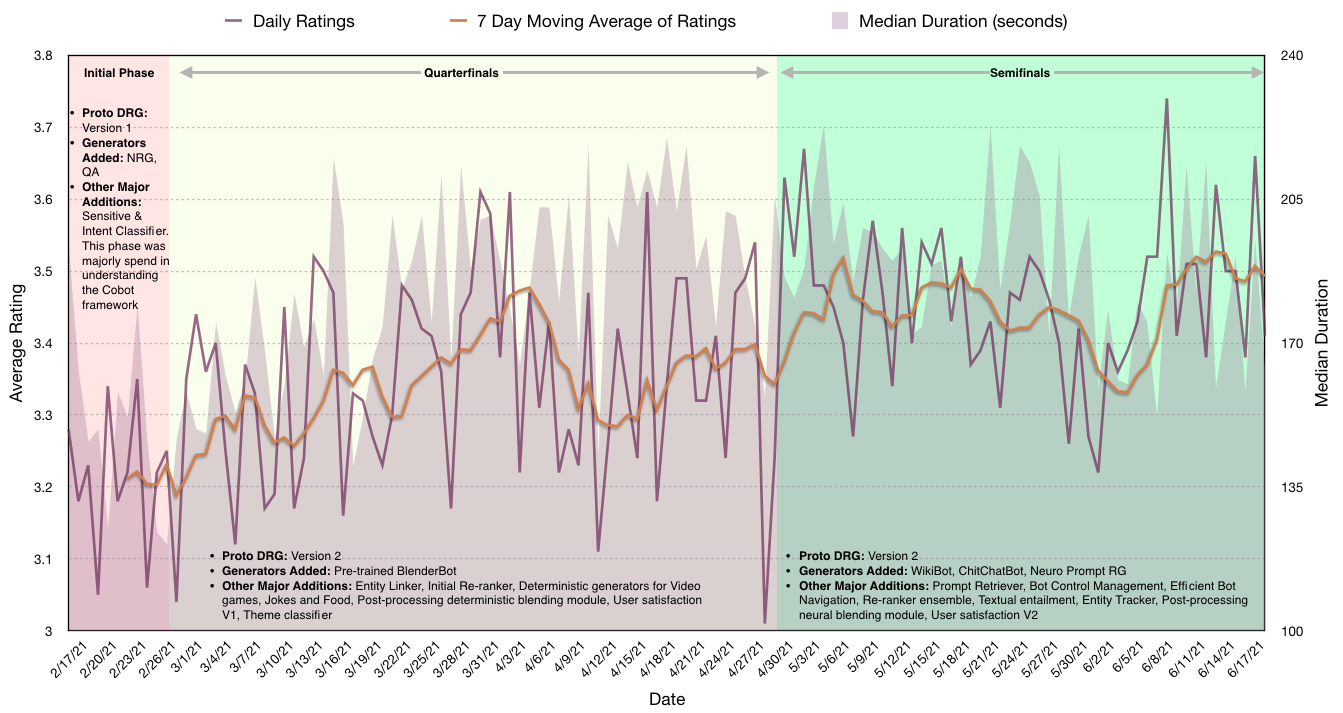}
    \caption{Proto Enhancement Timeline}
    \label{fig:proto_timeline}
\end{figure}

Throughout the competition, we received daily ratings from the users who interacted with the bot. Although the daily ratings tend to be volatile, overall we observe a positive trend. During the initial feedback phase our 7 day moving average of score varied between 3.15 and 3.23. With our constant improvements, the moving average varied between 3.23 and 3.48 during the quarterfinals. With the addition of the neural generators, improved maneuverability, ensemble of re-rankers and post-processors, the range of the moving average increased to lie between 3.32 and 3.52 in the semifinals. Figure \ref{fig:proto_timeline} illustrates an annotated timeline of the bot's performance throughout the competition.

\begin{figure}[h]
    \centering
    \includegraphics[scale=0.31]{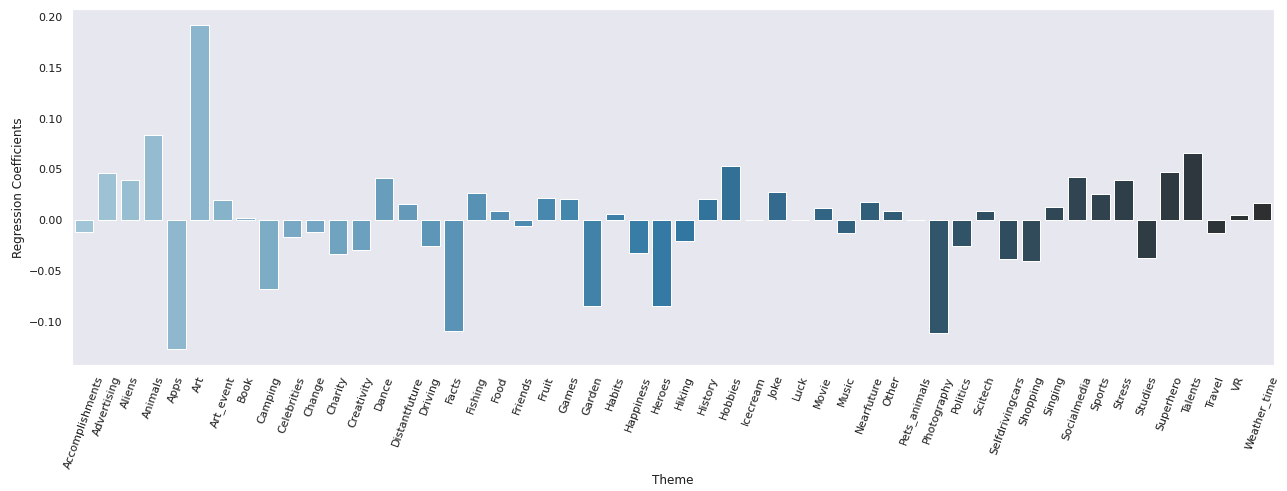}
    \caption{Coefficients of each theme count when used to predict ratings using OLS model.}
    \label{fig:theme_coeff}
\end{figure}

In order to analyze the impact of each theme in the conversations, we sampled all the conversations between 20th May and 20th June 2021, and fit an ordinary least square model, with the independent variable being the count of each theme (except Launch and Greeting as they are present in almost every conversation) in a conversation, and the dependent variable being the user rating. Figure \ref{fig:theme_coeff} illustrates the coefficients of each theme. We observe that themes Art, Talents, Superheroes, Hobbies, Stress, Social media, and Animals have the highest positive coefficients, whereas Apps, Facts, Heroes, Garden, Photography and Camping have high negative coefficients. 

\begin{figure}[h!]
  \centering
  \begin{minipage}{.6\textwidth}
    \includegraphics[width=\linewidth, scale=0.5]{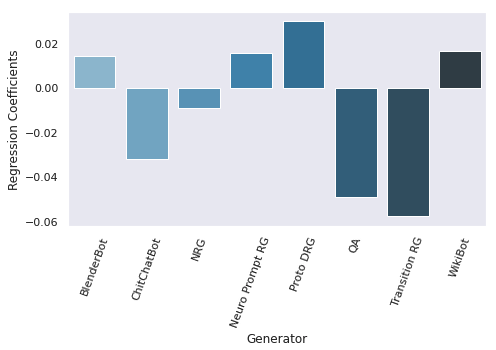}
    \caption{Coefficients of each generator count when \\used to predict ratings using OLS model.}
    \label{fig:generator_coeff}
  \end{minipage}%
%   \begin{minipage}{.45\textwidth}
%     \includegraphics[width=\linewidth, scale=0.5]{Alexa Prize Technical Article LaTeX/images/generaator_frequency_2.png}
%     \caption{Average number of turns by each generator, from April 1st onward.}
%     \label{fig:generator_freq}
%   \end{minipage}
    \begin{minipage}{.41\textwidth}
    \includegraphics[width=\linewidth, scale=0.5]{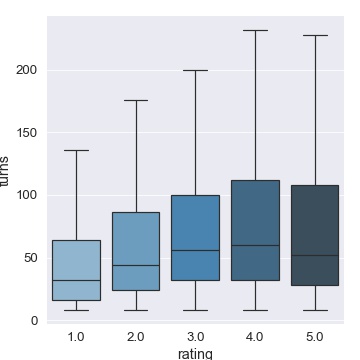}
    \caption{Relationship between ratings and number of turns in a conversation from April 1st till date.}
    \label{fig:turns_correl}
  \end{minipage}
%   \caption{Deterministic Generator}
%   \label{fig:deterministic_rg}
\end{figure}

\begin{figure}[h!]
    \centering
    \includegraphics[scale=0.4]{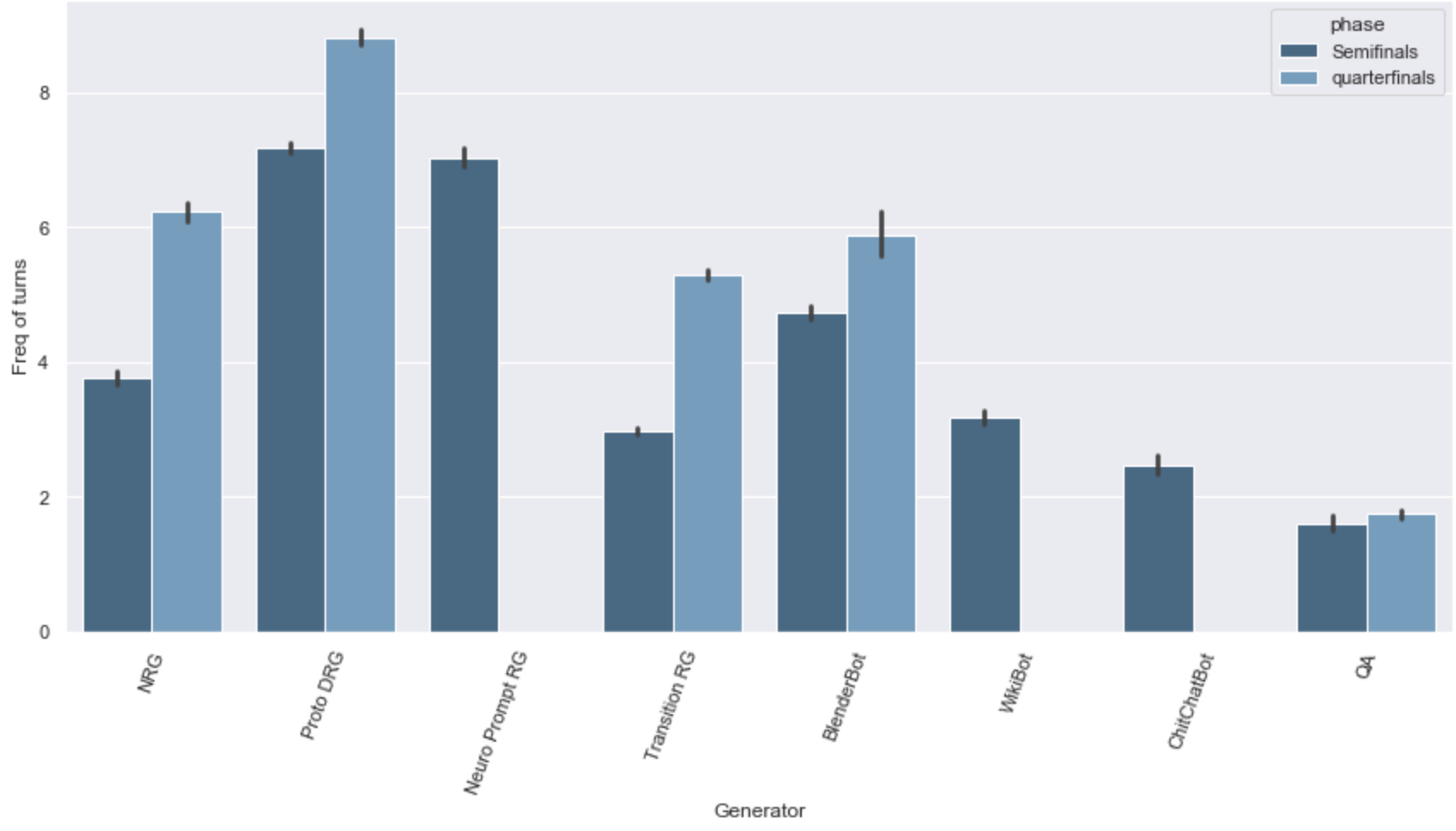}
    \caption{Comparison between frequency of turns generated by each generator during quarterfinals and semifinals.}
    \label{fig:generator_freq_quarterfinals_semifinals}
\end{figure}

\begin{figure}[h!]
    \centering
    \includegraphics[scale=0.47]{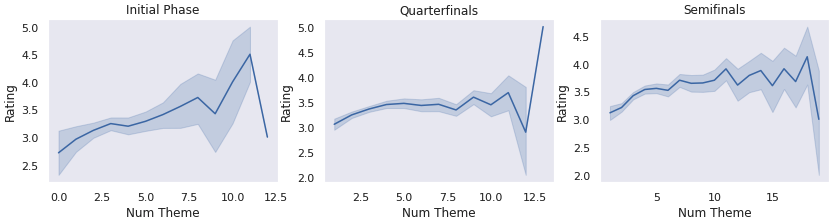}
    \caption{Relationship between ratings and number of themes discussed in a conversation for each phase of the competition}
    \label{fig:theme_correl}
\end{figure}

We also used ordinary least square method to understand how much each generator contribute to the ratings. Using the count of turns generated by each generator (except Launch and Invalid generators, as these generate only one turn of response) as the independent variables, and rating as the dependent variable, Figure \ref{fig:generator_coeff} illustrates the OLS coefficient for each generator. We observe that Proto DRG, Neuro Prompt RG, BlenderBot and WikiBot positively influence the ratings, whereas QA (EVI), Transition RG, ChitChatBot, and NRG negatively influence the ratings. We attribute the negative coefficient of ChitChatBot to the recency of its addition which resulted in a lack of sufficient data points for the regression model. The negative coefficient of NRG can be attributed to its mediocre response generation and abrupt theme switches, which leads to a bad user experience. We reason that the long Wikipedia based responses generated by QA generator might cause a bad user experience, hence the negative coefficient. Since the transition generator is used to transition between themes, a high count of responses generated by the transition generator either means a long engaging conversation, or a conversation where the user is not interested in the themes we introduce. Either way, we have observed longer conversations beyond a threshold to be correlated with lower ratings (Figure \ref{fig:theme_correl}), which explains the coefficients. In order to understand better the generators which generate most of the responses in a conversation, we plot and compare the frequency of turns generated per generator across all conversations in the quarterfinals and semifinals phase in Figure \ref{fig:generator_freq_quarterfinals_semifinals}. We observe that although Proto DRG generates most of the response in both the phases of the competition, the addition of the neural response generators in the semifinals, decrease the frequency of turns generated by the deterministic generator in favor of the neural generators.

% \begin{figure}[h]
%     \centering
%     \includegraphics[scale=0.4]{Alexa Prize Technical Article LaTeX/images/generator_Score_coef.png}
%     \caption{Coefficients of each generator count when used to predict ratings using OLS model.}
%     \label{fig:generator_coeff}
% \end{figure}

% \begin{figure}[h!]
%   \centering
%   \begin{minipage}{.42\textwidth}
%     \includegraphics[width=\linewidth, scale=0.5]{images/avg_turn_generator_quartefinals.png}
%     \caption{Frequency of turns generated \\by each generator in a conversation, dur-\\ing quarterfinals.}
%     \label{fig:generator_freq_quarterfinals}
%   \end{minipage}%
%     \begin{minipage}{.55\textwidth}
%     \includegraphics[width=\linewidth, scale=0.5]{Alexa Prize Technical Article LaTeX/images/avg_turn_generator_semifinals.png}
%     \caption{Frequency of turns generated by each generator in a conversation, during semifinals.}
%     \label{fig:generator_freq_semifinals}
%   \end{minipage}
% \end{figure}

\begin{figure}[h!]
    \centering
    \includegraphics[scale=0.47]{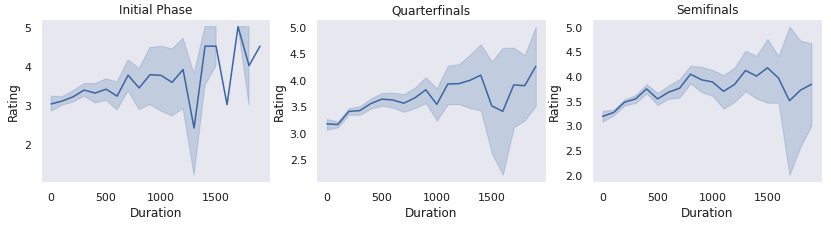}
    \caption{Relationship between ratings and duration of a conversation for each phase of the competition}
    \label{fig:duration_correl}
\end{figure}

In order to understand the relationship between conversation length and ratings, we analyzed the relationship between (i) the number of theme and ratings: Figure \ref{fig:theme_correl}, (ii) conversation duration and ratings: Figure \ref{fig:duration_correl}, and (iii) number of turns and ratings: Figure \ref{fig:turns_correl}. Almost all the plots suggest that an increase in ratings is positively correlated with conversation length, till a certain point, before which the ratings tend to go down. We attribute this phenomenon in longer conversations to repetitive themes, since almost all deterministic themes would have already been discussed.

% \begin{figure}[h]
%     \centering
%     \includegraphics[scale=0.3]{Alexa Prize Technical Article LaTeX/images/turn_ratings_all.jpeg}
%     \caption{Relationship between ratings and number of turns in a conversation from April $1^{st}$ till date}
%     \label{fig:turns_correl}
% \end{figure}

\section{Conclusion and Future Work}
Using concepts from information retrieval, natural language processing, machine learning, deep learning, and computational linguistics, Proto is an end-to-end conversational system made from a cocktail of multiple technologies. Leveraging a suite of natural language understanding modules, Proto can understand a user's intent, and leverage state-of-the-art neural and state-based deterministic generators, it can generate appropriate candidate responses, which are factual, empathetic, and colloquial. With the help of an ensemble of re-rankers and a post-processing module, Proto can select, augment and deliver the best response, given the current state of the conversation.

Although Proto is able to provide a positive and engaging experience to most of its users, there are scenarios where the system lacks, which is generally caused by contrasting results from one or more components of the bot. Recent research in deep learning has demonstrated the capabilities of end-to-end learning algorithms in circumventing such problems. Hence, in the future, we want to focus on coupling multiple modules together in order to leverage the end-to-end learning paradigm.

Since neural generators are trained on different datasets which have their own biases, efficiently controlling the generated neural response and post-processing it into a usable response that can be delivered to the user, is something we have found quite challenging. Also, leveraging multiple generators trained on multiple datasets often resulted in the generation of contradictory candidate responses. Hence, we believe that a significant amount of work needs to be done in creating training datasets that can reduce such contrasting personalities.

Content is king. Leveraging Wikidata and scraped online content, we are able to generate factual and engaging responses. However, the bot lacks the capability of establishing relationships between multiple facts and exploiting the relationships to dive deeper into a topic. We believe leveraging a knowledge graph-based approach can help the bot exploit the relationships between worldly facts, and generate deeper conversations. Overall, though there are several areas for continued improvement, Proto is a clear step forward in research on open-domain conversation.


% Use unnumbered third level headings for the acknowledgments \cite{labelsmoothing}. All
% acknowledgments go at the end of the paper. Do not include
% acknowledgments in the anonymized submission, only in the final paper.\cite{Jiang_2019}

% \section*{References}

% References follow the acknowledgments. Use unnumbered first-level
% heading for the references. Any choice of citation style is acceptable
% as long as you are consistent. It is permissible to reduce the font
% size to \verb+small+ (9 point) when listing the references. {\bf
%   Remember that you can use a ninth page as long as it contains
%   \emph{only} cited references.}
% \medskip
\bibliographystyle{ap_2021}
\bibliography{bibliography}

\begin{thebibliography}{10}

\bibitem{gopalakrishnan2019topical}
Gopalakrishnan, K., B.~Hedayatnia, Q.~Chen, et~al.
\newblock {Topical-Chat: Towards Knowledge-Grounded Open-Domain Conversations}.
\newblock In \emph{INTERSPEECH}. 2019.

\bibitem{myers2020conversational}
Myers, W., T.~Etchart, N.~Fulda.
\newblock Conversational scaffolding: An analogy-based approach to response
  prioritization in open-domain dialogs.
\newblock 2020.

\bibitem{convai2}
Dinan, E., V.~Logacheva, V.~Malykh, et~al.
\newblock The second conversational intelligence challenge (convai2).
\newblock \emph{CoRR}, abs/1902.00098, 2019.

\bibitem{devlin2019bert}
Devlin, J., M.-W. Chang, K.~Lee, et~al.
\newblock Bert: Pre-training of deep bidirectional transformers for language
  understanding, 2019.

\bibitem{paranjape2020neural}
Paranjape, A., A.~See, K.~Kenealy, et~al.
\newblock Neural generation meets real people: Towards emotionally engaging
  mixed-initiative conversations, 2020.

\bibitem{xu2020recipes}
Xu, J., D.~Ju, M.~Li, et~al.
\newblock Recipes for safety in open-domain chatbots, 2020.

\bibitem{clark-manning-2016-deep}
Clark, K., C.~D. Manning.
\newblock Deep reinforcement learning for mention-ranking coreference models.
\newblock In \emph{Proceedings of the 2016 Conference on Empirical Methods in
  Natural Language Processing}, pages 2256--2262. Association for Computational
  Linguistics, Austin, Texas, 2016.

\bibitem{clark-manning-2016-improving}
---.
\newblock Improving coreference resolution by learning entity-level distributed
  representations.
\newblock In \emph{Proceedings of the 54th Annual Meeting of the Association
  for Computational Linguistics (Volume 1: Long Papers)}, pages 643--653.
  Association for Computational Linguistics, Berlin, Germany, 2016.

\bibitem{Lee2017EndtoendNC}
Lee, K., L.~He, M.~Lewis, et~al.
\newblock End-to-end neural coreference resolution.
\newblock \emph{ArXiv}, abs/1707.07045, 2017.

\bibitem{erkan2004lexrank}
Erkan, G., D.~R. Radev.
\newblock Lexrank: Graph-based lexical centrality as salience in text
  summarization.
\newblock \emph{Journal of artificial intelligence research}, 22:457--479,
  2004.

\bibitem{rose2010automatic}
Rose, S., D.~Engel, N.~Cramer, et~al.
\newblock Automatic keyword extraction from individual documents.
\newblock \emph{Text mining: applications and theory}, 1:1--20, 2010.

\bibitem{liu2019roberta}
Liu, Y., M.~Ott, N.~Goyal, et~al.
\newblock Roberta: A robustly optimized bert pretraining approach, 2019.

\bibitem{ahmadvand2020concet}
Ahmadvand, A., H.~Sahijwani, J.~I. Choi, et~al.
\newblock Concet: Entity-aware topic classification for open-domain
  conversational agents, 2020.

\bibitem{reimers-2019-sentence-bert}
Reimers, N., I.~Gurevych.
\newblock Sentence-bert: Sentence embeddings using siamese bert-networks.
\newblock In \emph{Proceedings of the 2019 Conference on Empirical Methods in
  Natural Language Processing}. Association for Computational Linguistics,
  2019.

\bibitem{weizenbaum1966eliza}
Weizenbaum, J.
\newblock Eliza—a computer program for the study of natural language
  communication between man and machine.
\newblock \emph{Communications of the ACM}, 9(1):36--45, 1966.

\bibitem{Shawar2015ALICECT}
Shawar, B.~A., E.~Atwell.
\newblock Alice chatbot: Trials and outputs.
\newblock \emph{Computaci{\'o}n y Sistemas}, 19, 2015.

\bibitem{vaswani2017attention}
Vaswani, A., N.~Shazeer, N.~Parmar, et~al.
\newblock Attention is all you need, 2017.

\bibitem{Wolf2019HuggingFacesTS}
Wolf, T., L.~Debut, V.~Sanh, et~al.
\newblock Huggingface's transformers: State-of-the-art natural language
  processing.
\newblock \emph{ArXiv}, abs/1910.03771, 2019.

\bibitem{DBLP:journals/corr/abs-1901-08149}
Wolf, T., V.~Sanh, J.~Chaumond, et~al.
\newblock Transfertransfo: {A} transfer learning approach for neural network
  based conversational agents.
\newblock \emph{CoRR}, abs/1901.08149, 2019.

\bibitem{Zhang_2020}
Zhang, Y., S.~Sun, M.~Galley, et~al.
\newblock Dialogpt: Large-scale generative pre-training for conversational
  response generation.
\newblock \emph{Proceedings of the 58th Annual Meeting of the Association for
  Computational Linguistics: System Demonstrations}, 2020.

\bibitem{Shum_2018}
Shum, H.-y., X.-d. He, D.~Li.
\newblock From eliza to xiaoice: challenges and opportunities with social
  chatbots.
\newblock \emph{Frontiers of Information Technology \& Electronic Engineering},
  19(1):10–26, 2018.

\bibitem{Zhou_2020}
Zhou, L., J.~Gao, D.~Li, et~al.
\newblock The design and implementation of xiaoice, an empathetic social
  chatbot.
\newblock \emph{Computational Linguistics}, 46(1):53–93, 2020.

\bibitem{roller2020recipes}
Roller, S., E.~Dinan, N.~Goyal, et~al.
\newblock Recipes for building an open-domain chatbot, 2020.

\bibitem{hedayatnia2020policydriven}
Hedayatnia, B., K.~Gopalakrishnan, S.~Kim, et~al.
\newblock Policy-driven neural response generation for knowledge-grounded
  dialogue systems, 2020.

\bibitem{dinan2019wizard}
Dinan, E., S.~Roller, K.~Shuster, et~al.
\newblock Wizard of wikipedia: Knowledge-powered conversational agents, 2019.

\bibitem{rashkin2019empathetic}
Rashkin, H., E.~M. Smith, M.~Li, et~al.
\newblock Towards empathetic open-domain conversation models: a new benchmark
  and dataset, 2019.

\bibitem{zhang2018personalizing}
Zhang, S., E.~Dinan, J.~Urbanek, et~al.
\newblock Personalizing dialogue agents: I have a dog, do you have pets too?,
  2018.

\bibitem{smith2020together}
Smith, E.~M., M.~Williamson, K.~Shuster, et~al.
\newblock Can you put it all together: Evaluating conversational agents'
  ability to blend skills, 2020.

\bibitem{loshchilov2019decoupled}
Loshchilov, I., F.~Hutter.
\newblock Decoupled weight decay regularization, 2019.

\bibitem{holtzman2020curious}
Holtzman, A., J.~Buys, L.~Du, et~al.
\newblock The curious case of neural text degeneration, 2020.

\bibitem{li2017dailydialog}
Li, Y., H.~Su, X.~Shen, et~al.
\newblock Dailydialog: A manually labelled multi-turn dialogue dataset, 2017.

\bibitem{post2018clarity}
Post, M.
\newblock A call for clarity in reporting bleu scores, 2018.

\bibitem{gao2020dialogue}
Gao, X., Y.~Zhang, M.~Galley, et~al.
\newblock Dialogue response ranking training with large-scale human feedback
  data.
\newblock \emph{arXiv preprint arXiv:2009.06978}, 2020.

\bibitem{polyencoder}
Humeau, S., K.~Shuster, M.~Lachaux, et~al.
\newblock Real-time inference in multi-sentence tasks with deep pretrained
  transformers.
\newblock \emph{CoRR}, abs/1905.01969, 2019.

\bibitem{biencoder}
Mazar{\'{e}}, P., S.~Humeau, M.~Raison, et~al.
\newblock Training millions of personalized dialogue agents.
\newblock \emph{CoRR}, abs/1809.01984, 2018.

\bibitem{crossencoder}
Wolf, T., V.~Sanh, J.~Chaumond, et~al.
\newblock Transfertransfo: {A} transfer learning approach for neural network
  based conversational agents.
\newblock \emph{CoRR}, abs/1901.08149, 2019.

\bibitem{ubuntu}
Lowe, R., N.~Pow, I.~Serban, et~al.
\newblock The ubuntu dialogue corpus: {A} large dataset for research in
  unstructured multi-turn dialogue systems.
\newblock \emph{CoRR}, abs/1506.08909, 2015.

\bibitem{microreranker}
Gao, X., Y.~Zhang, M.~Galley, et~al.
\newblock Dialogue response ranking training with large-scale human feedback
  data.
\newblock \emph{CoRR}, abs/2009.06978, 2020.

\bibitem{Gardner2017AllenNLP}
Gardner, M., J.~Grus, M.~Neumann, et~al.
\newblock Allennlp: A deep semantic natural language processing platform.
\newblock 2017.

\bibitem{N18-1101}
Williams, A., N.~Nangia, S.~Bowman.
\newblock A broad-coverage challenge corpus for sentence understanding through
  inference.
\newblock In \emph{Proceedings of the 2018 Conference of the North American
  Chapter of the Association for Computational Linguistics: Human Language
  Technologies, Volume 1 (Long Papers)}, pages 1112--1122. Association for
  Computational Linguistics, 2018.

\end{thebibliography}
% \small

% [1] Alexander, J.A.\ \& Mozer, M.C.\ (1995) Template-based algorithms
% for connectionist rule extraction. In G.\ Tesauro, D.S.\ Touretzky and
% T.K.\ Leen (eds.), {\it Advances in Neural Information Processing
%   Systems 7}, pp.\ 609--616. Cambridge, MA: MIT Press.

% [2] Bower, J.M.\ \& Beeman, D.\ (1995) {\it The Book of GENESIS:
%   Exploring Realistic Neural Models with the GEneral NEural SImulation
%   System.}  New York: TELOS/Springer--Verlag.

% [3] Hasselmo, M.E., Schnell, E.\ \& Barkai, E.\ (1995) Dynamics of
% learning and recall at excitatory recurrent synapses and cholinergic
% modulation in rat hippocampal region CA3. {\it Journal of
%   Neuroscience} {\bf 15}(7):5249-5262.

\end{document}